\documentclass[]{fairmeta}

\usepackage{amsmath,amsfonts,bm}
\usepackage{xspace}
\usepackage{hyperref}
\usepackage{url}
\usepackage{subcaption}

\usepackage{colortbl}
\usepackage{xcolor}

\usepackage{graphicx}
\usepackage{algorithmic}
\usepackage{algorithm}
\usepackage{float}

\usepackage{booktabs}

\usepackage{enumitem}

\usepackage{amsthm}

\theoremstyle{plain}

\theoremstyle{definition}

\theoremstyle{remark}

\def\eqref#1{equation~\ref{#1}}

\def\1{\bm{1}}

\DeclareMathAlphabet{\mathsfit}{\encodingdefault}{\sfdefault}{m}{sl}
\SetMathAlphabet{\mathsfit}{bold}{\encodingdefault}{\sfdefault}{bx}{n}

\usepackage{subcaption}
\usepackage{wrapfig}
\usepackage[most]{tcolorbox}
\usepackage{xcolor}
\usepackage[table]{xcolor}
\usepackage{hyperref}
\usepackage{url}
\usepackage{multirow}
\usepackage{amsmath}
\usepackage{amsfonts}
\usepackage{graphicx}
\usepackage{listings}
\usepackage{xspace}
\usepackage{enumitem}
\usepackage{seqsplit}
\usepackage{fancyvrb}
\usepackage{fvextra}

\definecolor{softgreen}{RGB}{110, 160, 120}
\usepackage{svg}

\newtcolorbox{takeawaybox_basemodel}[1]{
    colback=orange!5!white,
    colframe=black,
    arc=5pt,
    outer arc=5pt,
    boxrule=0.8pt,
    left=5pt,
    right=5pt,
    top=4pt,
    bottom=4pt,
    fontupper=\small,
    enhanced,
    before upper={\textbf{#1 }} 
}

\newtcolorbox{promptbox}[1][]{
    colback=gray!5,
    colframe=gray!50,
    fonttitle=\bfseries\small,
    title=#1,
    breakable,
    left=4pt, right=4pt, top=4pt, bottom=4pt,
    fontupper=\small\ttfamily,
}

\usepackage{caption}
\usepackage{wrapfig}
\usepackage{booktabs}
\definecolor{earlyblue}{HTML}{88A2F1}
\definecolor{midgrey}{HTML}{fadcb4}
\definecolor{latered}{HTML}{EE9C88}
\definecolor{highlightgreen}{HTML}{80c66d}
\definecolor{highlightpurple}{HTML}{9b6d97}

\usepackage{arydshln}
\usepackage{xstring}
\newcommand{\deltaval}[1]{%
  \IfBeginWith{#1}{+}{%
    {\textcolor{highlightgreen}{\textit{(#1)}}}%
  }{%
    \IfBeginWith{#1}{-}{%
      {\textcolor{highlightpurple}{\textit{(#1)}}}%
    }{%
      {\textit{(#1)}}%
    }%
  }%
}
\usepackage{pifont}
\usepackage{epigraph}

\usepackage[export]{adjustbox}

\tcbset{
  finding style/.style={
    enhanced,
    colback=gray!10,           
    colframe=gray!60,         
    boxrule=0.8pt,             
    arc=2mm,                   
    left=1mm, right=1mm,       
    top=1mm, bottom=1mm,
    before skip=10pt, after skip=10pt,
    fontupper=\normalfont,     
    fonttitle=\normalfont\bfseries, 
  }
}

\newtcolorbox{findingbox}[1][]{finding style,#1}

\title{Cognitive Episodes in LLM Reasoning Traces Enable Interpretable Human Item Difficulty Prediction}

\author[*,1]{Chenguang Wang}
\author[*,2,3]{Ming Li}
\author[1]{Xinyue Zeng}
\author[4]{Zhuochun Li}
\author[2]{Hong Jiao}
\author[3]{Tianyi Zhou}
\author[1]{Dawei Zhou}

\renewcommand\affiliation[2][]{%
  \addtolist[#1]{#2}{\affiliationlist}{\affiliationformat}{~~~~}%
}

\affiliation[1]{Virginia Tech}
\affiliation[2]{University of Maryland}
\affiliation[3]{MBZUAI}
\affiliation[4]{University of Pittsburgh}

\contribution[*]{Co-first Author}

\abstract{

Predicting human item difficulty is a central challenge in educational assessment, where accurate estimates are critical for fairness and effective test construction.
Existing approaches typically rely on costly human calibration or item-level textual representations, offering limited interpretable evidence about the cognitive processes that make an item difficult.
We argue that item difficulty should be viewed not only as a property of item text, but also as an \textbf{observable consequence of the problem-solving burden induced by the item}.
Yet, this burden from human is difficult to observe directly or analyze at scale.
Reasoning-oriented large language models (LLMs) offer scalable process evidence through explicit reasoning traces, but raw traces are often lengthy and unstructured.
To organize this evidence, we introduce cognitive episodes, grounded in Schoenfeld's account of problem solving, which group trace segments into functional problem-solving states.
Building on this representation, we propose \textbf{\ours} (Episode to Difficulty), a framework that converts model-generated reasoning traces into episode sequences and extracts compact features capturing the scale, allocation, and transitions of reasoning effort.
These process features are combined with semantic item representations for human item difficulty prediction.
Experiments on four real-world datasets show that \ours consistently outperforms strong baselines, including fine-tuned SLMs, LLM in-context learning, and supervised LLM adaptation.
On the SAT-derived classification benchmarks, \ours achieves an 8.1\% average relative gain over supervised LLM fine-tuning baselines.
Further analyses show that harder items induce more effortful, iterative, and implementation-centered episode dynamics rather than merely longer responses.
These results establish cognitive episodes as a predictive and interpretable process representation for human item difficulty.
}

\metadata[Website]{\url{https://github.com/c-steve-wang/Epi2Diff}}

\authoremails{\email{\{cswang, dzhou\}@vt.edu}, 
\email{minglii@umd.edu}, 
\email{tianyi.zhou@mbzuai.ac.ae}}

\newcommand\ours{Epi2Diff\xspace}

\begin{document}

\maketitle

\section{Introduction}

Large-scale assessment is a core evaluation mechanism in domains such as college admissions, language testing, and medical licensure~\citep{hambleton1991fundamentals, eignor2013standards, hsu2018automated, alkhuzaey2021systematic}. In these settings, item difficulty prediction is important for constructing assessments with well-calibrated difficulty distributions and precise measurement along the ability scale~\citep{alkhuzaey2021systematic, parry2020ensuring}. Traditionally, item difficulty has been estimated from examinee responses using classical psychometric frameworks such as Classical Test Theory (CTT) and Item Response Theory (IRT) from responses collected during field testing~\citep{hsu2018automated, demars2010item, alkhuzaey2024text, benedetto2023quantitative}. Although effective, these approaches require substantial pretesting and are therefore costly, time-consuming, and difficult to scale.

In response to these limitations, prior work has explored automated approaches for predicting item difficulty.
Existing approaches mainly include feature-based methods and end-to-end fine-tuning methods~\citep{peters2025text,alkhuzaey2024text,benedetto2023quantitative}. Feature-based methods are generally more interpretable but depend on representation quality, whereas end-to-end fine-tuning methods, such as fine-tuning Small Language Models (SLMs) like BERT~\citep{devlin2019bert} or RoBERTa~\citep{liu2019roberta}, are more streamlined but often less interpretable and more vulnerable in small-data settings~\citep{peters2025text, rogoz2024unibucllm,benedetto2023quantitative, tack2024itec, xue2020predicting}.
This is particularly problematic in educational assessment, where both accuracy and interpretability are needed~\citep{messick1990validity,kane2013validating}.
More recently, LLMs, including reasoning-oriented models, have enabled new strategies for item difficulty prediction, including inference-time reasoning, fine-tuning, LLM-extracted features, reasoning- or sampling-augmented prediction, and uncertainty-based signals~\citep{rogoz2024unibucllm,duenas2024upn,razavi2025estimating,feng2025reasoning,zotos2024you,li2025item}.
While these methods show that LLMs can provide useful signals for item difficulty prediction, their effectiveness depends on how such signals are elicited and modeled. Further, they often rely on model-generated outputs as predictive cues without explicitly grounding them in interpretable evidence aligned with human psychometric item difficulty.
Meanwhile, a related but distinct line of work has operationalized difficulty in broader LLM settings, such as by stratifying data for easy-to-hard or cross-difficulty transfer or by scaling benchmark difficulty through required reasoning length~\citep{hase2024unreasonable,kordi2026revisiting, ding2025longreasonarena}.
However, these studies use difficulty mainly to analyze LLM behavior, rather than to transform reasoning processes into interpretable evidence for predicting human item difficulty.

Collectively, these limitations suggest that
\textbf{the key challenge for existing approaches lies not only in model capacity, but also in how human item difficulty is treated and modeled in prediction.}
Across much of the text-based item-difficulty literature, difficulty is treated as a single item-level outcome and predicted from item text, metadata, annotations, embeddings, or other model-derived representations~\citep{benedetto2023quantitative,alkhuzaey2024text}.
Although this formulation is psychometrically natural, it leaves open how to represent the cognitive and process-level factors that may be associated with observed difficulty.
Educational psychology suggests that such factors may include cognitive demand or problem-solving burden, such as coordination of operations, representation management, monitoring, verification, and effortful solution processes~\citep{sweller1988cognitive,sweller2011cognitive,noroozi2022scrutiny,schoenfeld2016learning}.

The challenge is that such cognitive demands underlying the item difficulty have historically been difficult to observe and model at scale. In assessment research, they are typically studied through indirect and labor-intensive methods such as think-aloud protocols and eye-tracking.
Recent reasoning models provide a new opportunity to operationalize this perspective because they often produce explicit reasoning traces for complex problem solving~\citep{guo2025deepseek, jaech2024openai, xiong2025self, muennighoff2025s1}.
In particular, they can expose process categories such as decomposition, implementation, revision, and verification that are largely hidden in response-only outputs from instruction-following LLMs and can be organized using cognitively grounded accounts of problem solving. Although these traces should not be treated as direct observations of human cognition, they can serve as scalable, model-generated proxies for solution processes.

However, translating this insight into an effective model is nontrivial, since model-generated reasoning traces are often long, redundant, and noisy. In item difficulty prediction, richer or higher-dimensional representations do not necessarily yield better predictions~\citep{fulari2024utilizing,tack2024itec}. Directly incorporating raw reasoning traces may therefore introduce irrelevant information and even degrade performance. Accordingly, the central question of this work is: \textbf{\emph{How can explicit model reasoning processes be transformed into a structured and informative representation for modeling item difficulty?}}

To tackle this challenge, we advance item difficulty prediction by augmenting semantic item representations with cognitive episode signals derived from model-generated reasoning traces.
Specifically, \textbf{\ours} (Episode to Difficulty) converts model-generated reasoning traces into sentence-level cognitive episode sequences using a cognitively grounded episode taxonomy~\citep{schoenfeld2014mathematical, schoenfeld2016learning, li2025understanding, li2025schoenfeld}.
These sequences provide a structured and interpretable view of reasoning dynamics. From these sequences, we derive compact process features that capture the scale, composition, and organization of the reasoning process, and combine them with semantic embeddings of item content to form the final representation for downstream difficulty prediction.
This representation is valuable not only for improving predictive accuracy but also for supporting interpretable analyses of item difficulty by identifying episode types and transition patterns associated with greater difficulty, such as prolonged implementation, repeated verification, or inefficient local transitions.
Experiments on four datasets spanning SAT, Cambridge English Qualifications, and USMLE show that our method consistently outperforms various baselines, including fine-tuning small encoder-only language models, LLM In-Context Learning, and LLM Supervised Fine-tuning with full parameters or LoRA \citep{hu2022lora}.
On the real-world SAT-derived classification benchmarks, \ours achieves an 8.1\% average relative gain over LLM supervised fine-tuning baselines, while also improving regression performance on Cambridge English Qualifications and USMLE.

\begin{figure}
    \centering
    \includegraphics[width=1\linewidth]{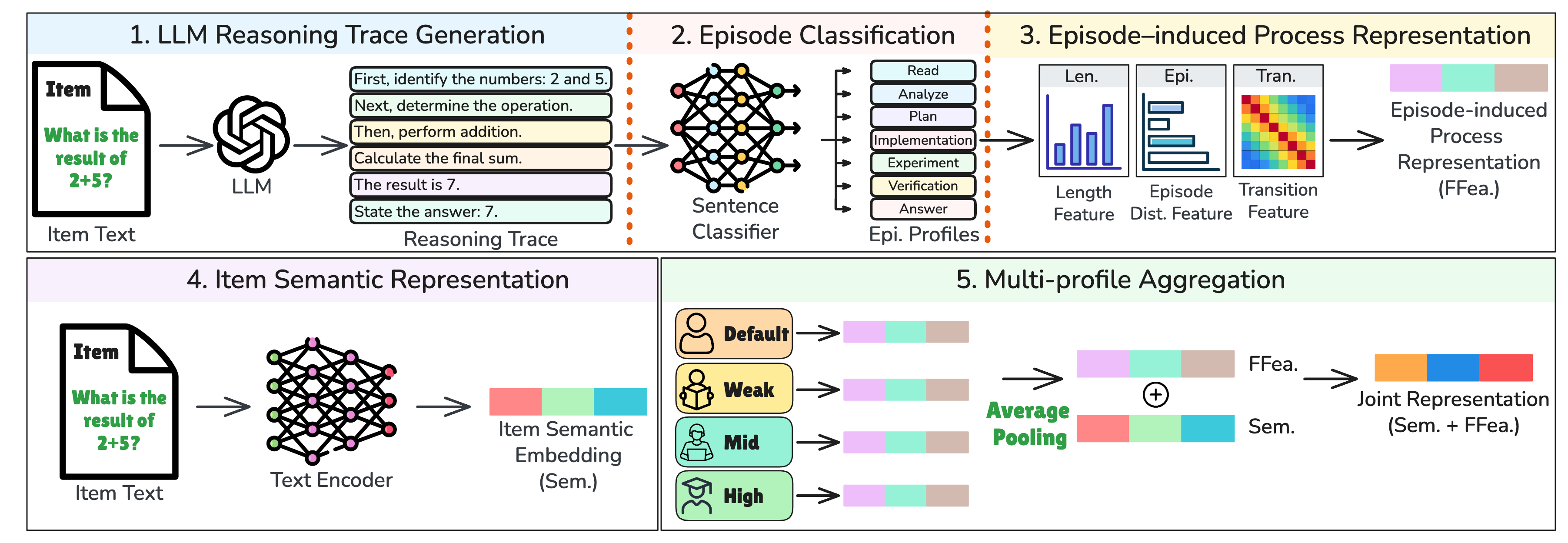}
    \caption{\textbf{Overview of Epi2Diff.} The framework generates reasoning traces with reasoning-oriented LLMs, maps them to sentence-level episode profiles, builds episode-induced process representations from length, episode distribution, and transition features, and combines them with item semantic embeddings through multi-profile aggregation for item difficulty prediction.}
    \label{fig:main_fig}
\end{figure}

Therefore, our contributions are threefold:

\begin{itemize}




    \item We extend human item difficulty prediction beyond text representations by \textbf{modeling difficulty as an observable consequence of item-induced problem-solving burden}, using model-generated reasoning traces as scalable proxy evidence.

    \item We develop a cognitively grounded episode representation that captures the scale, allocation, and flow of reasoning effort, enabling both accurate prediction and process-based interpretation.

    \item Across four real-world assessment benchmarks, \ours consistently outperforms strong baselines in classification and regression; further analyses show that harder items elicit more effortful, implementation-centered, and iterative reasoning dynamics rather than merely longer traces.
\end{itemize}

\section{Method}

\subsection{Preliminary}

\subsubsection{Schoenfeld’s Episode Theory}

Schoenfeld’s Episode Theory~\citep{schoenfeld2014mathematical} models the problem-solving process as a temporally ordered sequence of functional episodes.
Schoenfeld’s framework comprises seven episodes: six original episodes, \textit{Read}, \textit{Analyze}, \textit{Plan}, \textit{Implement}, \textit{Explore}, and \textit{Verify}, plus the later-added \textit{Monitor} episode. This framework is well-suited to our setting because it provides a cognitively grounded vocabulary for characterizing how problem-solving burden is distributed across different stages of long reasoning traces. Following ThinkARM~\citep{li2025schoenfeld}, we add an \textit{Answer} episode to capture final answer commitment, which is often distinct from preceding verification and monitoring. We therefore use an eight-way taxonomy consisting of \textit{Read}, \textit{Analyze}, \textit{Plan}, \textit{Implement}, \textit{Explore}, \textit{Verify}, \textit{Monitor}, and \textit{Answer}.

\subsubsection{Problem Formulation}

Let $\mathcal{D}=\{(x_i, y_i)\}_{i=1}^{N}$ denote a dataset of assessment items, where $x_i \in \mathcal{X}$ is a raw assessment item and $y_i \in \mathcal{Y}$ is its ground-truth difficulty label.
Depending on the label form, we consider either classification with $\mathcal{Y}=\{\text{Easy},\text{Medium},\text{Hard}\}$ or regression with $\mathcal{Y}\subseteq\mathbb{R}$.

For each item $x_i$, we query a reasoning-oriented LLM under a set of prompting conditions
$\mathcal{P}=\{p_m\}_{m=1}^{M}$. Under condition $p_m$, the rollout operator $\Phi$
produces a reasoning rollout
$r_i^{(m)}=\Phi(x_i,p_m)=\bigl(\tau_i^{(m)},\hat{a}_i^{(m)}\bigr)$, and the resulting
rollout set is $\mathcal{R}_i=\{r_i^{(m)}\}_{m=1}^{M}$. We then construct an item
representation from $x_i$ and $\mathcal{R}_i$ and predict difficulty via
\begin{equation}
\hat{y}_i = f\bigl(S(x_i,\mathcal{R}_i)\bigr), \qquad y_i \in \mathcal{Y},
\end{equation}
where $S$ maps the item and its associated reasoning rollouts into a representation
combining item-side semantic information and process signals derived from the reasoning
traces, and $f$ is a classifier or regressor depending on the label space.

The model is trained by minimizing
\begin{equation}
\min_{\theta}\frac{1}{N}\sum_{i=1}^{N}\ell(\hat{y}_i,y_i),
\end{equation}
where $\ell$ is the task-specific loss, e.g., cross-entropy for classification and mean
squared error for regression.

\subsection{Sentence-Level Episode Profiling from Reasoning Traces}

To support sentence-level episode profiling, we build a labeled corpus from the ThinkARM reasoning-trace annotations. ThinkARM is constructed from reasoning traces collected on Omni-MATH~\citep{gao2024omni} and annotated at the sentence level under the episode taxonomy introduced earlier. In this work, we use the subset generated by five representative reasoning models, including DeepSeek-R1 \citep{guo2025deepseek}, R1-Distill-Qwen3-32B \citep{guo2025deepseek}, QwQ-32B \citep{qwq32b}, Qwen3-32B \citep{yang2025qwen3}, and Phi4-Reasoning \citep{abdin2024phi}. Each reasoning trace is segmented into sentences using a standard sentence-splitting pipeline, and each sentence is paired with its corresponding episode label. The annotation resources include both large-scale GPT-5 annotations and a human-verified gold set of 7,067 annotated sentences. We use the GPT-5-annotated data for training and validation, and reserve the human-annotated gold set for final testing. Since our goal is sentence-level episode prediction, we flatten the reasoning traces into a unified pool of sentence, episode pairs. After basic cleaning to remove invalid samples, reduce label inconsistencies, and deduplicate repeated pairs, the final corpus contains 231,913 labeled sentence, episode pairs for training and validation.

We train a RoBERTa-based sentence classifier on the cleaned corpus for eight-way episode prediction. On the human-annotated gold test set, the classifier achieves an accuracy of $0.799$, a macro-F1 of $0.761$, and a micro-F1 of $0.799$. We then use the trained classifier as an automatic episode labeler in our following method. For each reasoning trace, we apply the same sentence segmentation pipeline and predict an episode label for every sentence, thereby inducing an episode-level representation for each item.

\subsection{Item Representation}

Prior work has shown that reasoning traces exhibit structured episode-level dynamics, suggesting that the problem-solving burden of an item may be reflected in both episode-wise properties and inter-episode transitions. Building on this insight, we construct an episode-based representation from labeled reasoning traces to model such burden and improve item difficulty prediction. To determine the most effective representation construction, we systematically explored episode-statistic features at multiple levels, ranging from basic episode distributions to multi-hop transition patterns, and empirically evaluated their individual contributions and combinations in downstream item difficulty prediction.

Based on this feature search and performance comparison, we adopt a two-part item representation: (1) an item semantic representation, which encodes the item content; and (2) episode-induced process representation for capturing the problem-solving burden reflected in reasoning traces. Concretely, the episode-induced process representation combines three complementary groups of features: length features capturing the overall scale of the reasoning process, episode-distribution features characterizing its composition across episode types, and transition features encoding its local sequential organization through inter-episode transitions.

\subsubsection{Episode-induced process representation}

For each item, we represent its reasoning trace as a labeled episode sequence over ordered sentence-level units and derive the episode-induced process representation from it.

\paragraph{Length features}
We first construct three length features for each item from its labeled reasoning trace, corresponding to the total token length of the trace, the token length of the thinking part, and the token length of the answer part.
Here, the \texttt{think}/\texttt{answer} distinction follows the structure of the model response, where the explicit reasoning-trace portion is separated from the final-answer portion, and all lengths are measured in tokens.
We denote these three quantities by \(l_{i,\mathrm{total}}\), \(l_{i,\mathrm{think}}\), and \(l_{i,\mathrm{answer}}\), and group them into a 3-dimensional length vector \(\mathbf{L}_i = [l_{i,\mathrm{total}}; l_{i,\mathrm{think}}; l_{i,\mathrm{answer}}] \in \mathbb{R}^3\). The resulting vector captures the overall scale of the reasoning process while distinguishing reasoning effort from final answer production.

\paragraph{Episode-distribution features}
We next characterize how the reasoning budget is distributed across episode types. For each episode \(e \in \mathcal{E}\), we compute an episode token count \(c_{i,e}\) as the total token length of all labeled sentence-level units assigned to episode \(e\). We then normalize these episode counts to obtain episode ratios:
\begin{equation}
r_{i,e} = \frac{c_{i,e}}{\sum_{e' \in \mathcal{E}} c_{i,e'}}.
\end{equation}
Based on these quantities, we form an episode-count vector \(\mathbf{C}_i = [c_{i,e}]_{e \in \mathcal{E}} \in \mathbb{R}^{|\mathcal{E}|}\) and an episode-ratio vector \(\mathbf{R}_i = [r_{i,e}]_{e \in \mathcal{E}} \in \mathbb{R}^{|\mathcal{E}|}\).
The count vector preserves the absolute amount of trace content allocated to each episode, whereas the ratio vector captures the relative allocation of reasoning effort after normalizing for total episode token mass.
The two vectors jointly preserve the overall reasoning scale and enable comparison across traces.

\paragraph{Transition features}
Finally, we characterize reasoning flow through episode transitions. Rather than treating a trace as an unordered collection of episode statistics, we record how often it moves from one episode type to the next along the labeled sequence. For any pair of episodes \((e,e') \in \mathcal{E}\times\mathcal{E}\), we define a transition count \(a_{i,e,e'}\) as the number of times episode \(e\) is immediately followed by episode \(e'\). Collecting all such counts yields an episode transition matrix \(\mathbf{A}_i = [a_{i,e,e'}]_{e,e' \in \mathcal{E}} \in \mathbb{R}^{|\mathcal{E}|\times|\mathcal{E}|}\). In our setting, \(|\mathcal{E}|=8\), so \(\mathbf{A}_i\) is an \(8\times 8\) matrix whose entries record raw adjacent transition counts.

This matrix captures local sequential organization that is not available from token totals alone. For example, two traces may allocate similar reasoning effort across episode types but still differ substantially in how they move through them over time. To combine transition features with the other process feature groups, we flatten \(\mathbf{A}_i\) into a 64-dimensional vector and concatenate it with the length, episode-count, and episode-ratio features. The resulting episode-induced process representation is therefore \(\mathbf{p}_i = [\mathbf{L}_i; \mathbf{C}_i; \mathbf{R}_i; \mathrm{vec}(\mathbf{A}_i)] \in \mathbb{R}^{83}\), where the 83 dimensions consist of 3 length features, 8 episode token counts, 8 episode token ratios, and 64 transition counts.

\subsubsection{Item Semantic Representation and Joint Representation}

\paragraph{Item semantic representation.}
Episode-induced process representations characterize the solution processes elicited from the selected reasoning models, but they do not directly encode the semantic content of the item itself.
We therefore complement \(\mathbf{p}_i\) with an item semantic representation obtained from a Sentence-BERT encoder \citep{reimers2019sentence}. Specifically, we encode the item text into a contextual embedding \(\mathbf{h}_i \in \mathbb{R}^{d}\), which captures its semantic meaning in a dense continuous space. This embedding provides information about the content and formulation of the item that is not directly reflected in reasoning-trace statistics.

\paragraph{Joint representation.}
We then combine the semantic and process components to form the joint representation, denoted by \(\mathbf{u}_i = [\mathbf{h}_i; \mathbf{p}_i]\). The item semantic representation \(\mathbf{h}_i\) captures the content of the item, whereas the Episode-induced process representation \(\mathbf{p}_i\) captures the reasoning burden and structural dynamics elicited by it. The resulting joint representation, therefore, integrates both what the item is about and how it is solved.

\subsection{Item Difficulty Prediction}

Given an item, we further construct its final representation by aggregating reasoning processes induced under multiple proficiency-conditioned model profiles \citep{yuan2026towards}. Inspired by psychological assessment frameworks that consider different ability levels, we use direct, weak, medium, and strong prompting conditions to elicit diverse solution processes from the reasoning model. These profiles are prompting conditions and should not be interpreted as validated simulations of human students or ability levels.
Let \(s \in \mathcal{S}\) index the model profile, where \(\mathcal{S} = \{\text{direct}, \text{weak}, \text{medium}, \text{strong}\}\). For each profile \(s\), we obtain a reasoning trace \(\tau_i^{(s)}\) and map it to an episode-induced process representation \(\mathbf{p}_i^{(s)}\). We then average these process representations across model profiles:
\begin{equation}
\bar{\mathbf{p}}_i = \frac{1}{|\mathcal{S}|} \sum_{s \in \mathcal{S}} \mathbf{p}_i^{(s)}.
\end{equation}
Since the underlying item content is identical across model profiles, the semantic representation \(\mathbf{h}_i\) is kept fixed and is not updated during this stage. The final item representation is then formed as \(\mathbf{u}_i = [\mathbf{h}_i; \bar{\mathbf{p}}_i]\).

We then train supervised predictors on top of the final item representation for difficulty estimation.
All predictors are trained and evaluated under the same train/validation/test split protocol, and preprocessing statistics are computed using only the training split to avoid data leakage.
For categorical difficulty estimation, we instantiate the predictor as a classifier; for continuous difficulty estimation, we instantiate it as a regressor. To evaluate the robustness of the proposed representation across different inductive biases, we experiment with both linear and tree-based predictors. For classification, we use Logistic Regression, Random Forest, and XGBoost. For regression, we use Ridge, Random Forest, and XGBoost. For scale-sensitive linear models, the input representation is standardized using training-split statistics, i.e., \(\tilde{\mathbf{u}}_i = (\mathbf{u}_i - \boldsymbol{\mu}) / \boldsymbol{\sigma}\), while tree-based models are trained on the original feature scale. This formulation allows us to evaluate whether the proposed episode-informed representation supports both linear and non-linear predictors under a unified setup.
For each dataset, we select the downstream predictor from the candidate set above based on validation performance. Notably, across all datasets, XGBoost is consistently chosen under this protocol, so the reported results for our method are all obtained with XGBoost in the following section.

\section{Experiments and Results}

\subsection{Experiment Setup}


\subsubsection{Datasets}
\paragraph{Datasets.}
We evaluate on four educational benchmarks spanning medical, English reading-comprehension, verbal reasoning, and mathematical reasoning domains.
\textbf{(1) USMLE.} The USMLE dataset~\citep{yaneva2024findings} contains 667 multiple-choice items from a high-stakes medical assessment setting, with difficulty labels derived from field-test data collected from more than 300 medical students per item. Following the original setup, we use the provided linearly transformed proportion-correct scores in $[0,1.3]$ as continuous difficulty labels, where higher values indicate more difficult items.
\textbf{(2) Cambridge.} The Cambridge dataset~\citep{cambridge_reading_comprehension} contains 120 reading passages and 793 multiple-choice questions for English reading comprehension, with rescaled IRT $b$-parameters in $[0,100]$ as continuous difficulty labels.
\textbf{(3) SAT Reading \& Writing.} This dataset is constructed from the SAT Question Bank\footnote{\url{https://satsuitequestionbank.collegeboard.org/}} after excluding figure-dependent items. It contains 1,338 verbal reasoning questions covering reading comprehension, writing, and language conventions, with three discrete labels: Easy, Medium, and Hard.
\textbf{(4) SAT Math.} This dataset is also collected from the SAT Question Bank and contains 1,075 mathematics questions covering algebra, geometry, data analysis, and multi-step problem solving, using the same Easy, Medium, and Hard labels.
Detailed train--test split sizes and protocols are provided in Appendix~\ref{app:dataset_splits}.

\subsubsection{Models and Baselines}
Our approach begins by leveraging two strong reasoning-oriented LLMs, \textbf{QwQ-32B} \citep{qwq32b} and \textbf{Qwen3-32B} \citep{yang2025qwen3}, in reasoning mode, to produce explicit reasoning traces for each item. These generated traces serve as intermediate supervision signals and are subsequently used as the foundation for downstream difficulty prediction.

To provide a fair and comprehensive comparison against mainstream item difficulty prediction methods, we evaluate baselines from \textbf{five} representative paradigms, spanning both conventional small-model approaches and recent LLM-based strategies.
\textbf{(1) SLM fine-tuning:} We fine-tune encoder-based small language models, including BERT \citep{devlin2019bert}, RoBERTa \citep{liu2019roberta}, ModernBERT \citep{warner2024smarterbetterfasterlonger}, and ELECTRA \citep{clark2020electrapretrainingtextencoders}, as standard supervised baselines for difficulty prediction.
\textbf{(2) LLM zero-shot inference:} we directly prompt frontier LLMs, including GPT-4o \citep{openai2024gpt4technicalreport}, GPT-5 \citep{singh2025openaigpt5card}, QwQ-32B \citep{qwq32b}, and Qwen3-32B \citep{yang2025qwen3}, to predict item difficulty without task-specific training.
\textbf{(3) LLM in-context learning:} we further evaluate the same set of LLMs under few-shot prompting with different shot settings, in order to test whether exemplar-based conditioning improves prediction quality, including GPT-4o \citep{openai2024gpt4technicalreport}, GPT-5 \citep{singh2025openaigpt5card}, QwQ-32B \citep{qwq32b}, and Qwen3-32B \citep{yang2025qwen3}.
\textbf{(4) LLM supervised fine-tuning:} we conduct full-parameter supervised fine-tuning on a group of compact open-source decoder models, including Qwen2.5-3B \citep{qwen2.5}, Qwen3-4B \citep{yang2025qwen3}, Phi-3.5-mini \citep{abdin2024phi3technicalreporthighly}, Phi-4-mini \citep{abdin2024phi}, and Llama-3.2-3B \citep{grattafiori2024llama3herdmodels}.
\textbf{(5) Parameter-efficient fine-tuning:} on the same model set, we additionally implement \textbf{LoRA}-based adaptation as a more efficient alternative to full fine-tuning \citep{hu2022lora}, using a fixed rank of $r=16$ for all experiments. Together, these baselines cover a broad methodological spectrum, allowing us to compare our approach against both classical supervised models and modern prompting- and fine-tuning-based LLM methods under a unified experimental setup.

\paragraph{Evaluation metrics.}
For classification benchmarks, we report accuracy and macro-F1, where macro-F1 averages class-wise F1 scores equally across difficulty classes. For regression benchmarks, we report RMSE and $R^2$. Lower RMSE and higher accuracy, macro-F1, and $R^2$ indicate better performance. Detailed definition of evaluation metrics is provided in Appendix~\ref{appdx:detail_exp}.

\subsection{Main Results}

\begin{table*}[t]
    \centering
    \scriptsize
    \renewcommand{\arraystretch}{1.15}
    \setlength{\tabcolsep}{3pt}
    \begin{tabular*}{0.9\textwidth}{@{\extracolsep{\fill}}lcccccccc@{}}
    \toprule
    \textbf{Method}
    & \multicolumn{2}{c}{\textbf{SAT Math}}
    & \multicolumn{2}{c}{\textbf{SAT Reading}}
    & \multicolumn{2}{c}{\textbf{Cambridge}}
    & \multicolumn{2}{c}{\textbf{USMLE}} \\
    \cmidrule{2-3}
    \cmidrule{4-5}
    \cmidrule{6-7}
    \cmidrule{8-9}
    & \textbf{ACC $\uparrow$} & \textbf{F1 $\uparrow$}
    & \textbf{ACC $\uparrow$} & \textbf{F1 $\uparrow$}
    & \textbf{RMSE $\downarrow$} & \textbf{$R^2 \uparrow$}
    & \textbf{RMSE $\downarrow$} & \textbf{$R^2 \uparrow$} \\
    \midrule
    \multicolumn{9}{c}{\emph{SLM Fine-tuning}} \\[-2pt]
    \midrule
    BERT \citep{devlin2019bert} & 0.619 & 0.622 & 0.619 & 0.617 & 8.215 & 0.296 & 0.302 & 0.052 \\
    RoBERTa \citep{liu2019roberta} & 0.601 & 0.602 & 0.623 & 0.618 & 8.311 & 0.280 & 0.298 & 0.076 \\
    ModernBERT \citep{warner2024smarterbetterfasterlonger} & 0.637 & 0.643 & 0.601 & 0.598 & 8.200 & 0.299 & 0.306 & 0.026 \\
    ELECTRA \citep{clark2020electrapretrainingtextencoders} & 0.609 & 0.604 & 0.623 & 0.622 & 8.277 & 0.285 & 0.304 & 0.043 \\
    \midrule
    \multicolumn{9}{c}{\emph{LLM Zero-shot} Inference and In-Context Learning} \\[-2pt]
    \midrule
    GPT-4o (0-shot) \citep{openai2024gpt4technicalreport} & 0.447 & 0.350 & 0.478 & 0.397 & 13.693 & -0.956 & 0.347 & -0.246 \\
    GPT-5 (0-shot) \citep{singh2025openaigpt5card} & 0.349 & 0.188 & 0.381 & 0.225 & 33.657 & -10.815 & 0.365 & -0.380 \\
    QwQ-32B (0-shot) \citep{qwq32b} & 0.405 & 0.282 & 0.477 & 0.384 & 32.022 & -9.694 & 0.349 & -0.165 \\
    Qwen3-32B (0-shot) \citep{yang2025qwen3} & 0.449 & 0.304 & 0.440 & 0.349 & 19.376 & -2.916 & 0.389 & -0.509 \\
    \midrule
    GPT-4o (3-shot) \citep{openai2024gpt4technicalreport} & 0.391 & 0.376 & 0.246 & 0.255 & 24.518 & -5.217 & 0.358 & -0.357 \\
    GPT-5 (3-shot) \citep{singh2025openaigpt5card} & 0.358 & 0.212 & 0.384 & 0.230 & 23.026 & -4.529 & 0.350 & -0.269 \\
    QwQ-32B (3-shot) \citep{qwq32b} & 0.321 & 0.321 & 0.011 & 0.020 & 26.787 & -6.102 & 0.399 & -0.569 \\
    Qwen3-32B (3-shot) \citep{yang2025qwen3} & 0.405 & 0.359 & 0.299 & 0.335 & 16.371 & -1.853 & 0.393 & -0.559 \\
    \midrule
    GPT-4o (5-shot) \citep{openai2024gpt4technicalreport} & 0.381 & 0.378 & 0.291 & 0.301 & 16.181 & -1.758 & 0.353 & -0.272 \\
    GPT-5 (5-shot) \citep{singh2025openaigpt5card} & 0.349 & 0.202 & 0.381 & 0.225 & 19.540 & -2.964 & 0.351 & -0.277 \\
    QwQ-32B (5-shot) \citep{qwq32b} & 0.237 & 0.264 & 0.022 & 0.042 & 25.537 & -5.752 & 0.402 & -0.647 \\
    Qwen3-32B (5-shot) \citep{yang2025qwen3} & 0.391 & 0.374 & 0.235 & 0.285 & 15.122 & -1.500 & 0.513 & -2.084 \\

    \midrule
    \multicolumn{9}{c}{\emph{LLM Supervised Fine-tuning (Full Parameter and LoRA)}} \\[-2pt]
    \midrule
    Qwen2.5-3B (full) \citep{qwen2.5} & 0.660 & 0.667 & 0.578 & 0.573 & 10.227 & -0.107 & 0.478 & -1.368 \\
    Qwen3-4B (full) \citep{yang2025qwen3} & 0.698 & 0.705 & 0.619 & 0.618 & 10.242 & -0.386 & 0.478 & -1.368 \\
    Phi-3.5-mini (full) \citep{abdin2024phi3technicalreporthighly} & 0.674 & 0.681 & 0.556 & 0.540 & 14.045 & -1.152 & 0.485 & -1.441 \\
    Phi-4-mini (full) \citep{abdin2024phi} & 0.684 & 0.691 & 0.619 & 0.615 & 11.055 & -0.258 & 0.459 & -1.188 \\
    Llama3.2-3B (full) \citep{grattafiori2024llama3herdmodels} & 0.679 & 0.687 & 0.601 & 0.597 & 9.380 & 0.086 & 0.460 & -1.196 \\

    \midrule
    Qwen2.5-3B (LoRA) \citep{qwen2.5} & 0.628 & 0.611 & 0.608 & 0.602 & 13.419 & -1.004 & 0.371 & -0.431 \\
    Qwen3-4B (LoRA) \citep{yang2025qwen3} & 0.628 & 0.629 & 0.590 & 0.585 & 10.524 & -0.228 & 0.414 & -0.775 \\
    Phi-3.5-mini (LoRA) \citep{abdin2024phi3technicalreporthighly} & 0.637 & 0.631 & 0.586 & 0.587 & 12.779 & -0.985 & 0.449 & -1.093 \\
    Phi-4-mini (LoRA) \citep{abdin2024phi} & 0.647 & 0.648 & 0.541 & 0.536 & 10.523 & -0.221 & 0.469 & -1.284 \\
    Llama3.2-3B (LoRA) \citep{grattafiori2024llama3herdmodels} & 0.595 & 0.591 & 0.590 & 0.590 & 9.117 & 0.076 & 0.450 & -1.097 \\
    \midrule
    \textbf{\ours (ours)} & \textbf{0.730} & \textbf{0.728} & \textbf{0.631} & \textbf{0.626} & \textbf{7.612} & \textbf{0.396} & \textbf{0.291} & \textbf{0.121} \\
    \bottomrule
    \end{tabular*}
    \caption{
\textbf{Main results} on four item difficulty prediction benchmarks: SAT Math, SAT Reading, Cambridge, and USMLE. We compare \ours with representative baselines from SLM fine-tuning, zero-shot and in-context LLM prompting, and supervised LLM adaptation under both full-parameter and LoRA settings. We report Accuracy and macro-F1 on SAT Math and SAT Reading, and RMSE and $R^2$ on Cambridge and USMLE. Best results are shown in bold.
}
    \label{tab:main_result}
\end{table*}

Table~\ref{tab:main_result} reports results on four test sets spanning two classification datasets (SAT Math and SAT Reading \& Writing) and two regression datasets (Cambridge and USMLE). Overall, our method achieves the best performance across all four test sets and all reported metrics, suggesting that structured representations derived from model-generated reasoning traces provide more informative signals for item difficulty prediction than conventional approaches.

For classification, our method performs best on both SAT datasets, with especially clear gains on SAT Math. This suggests that reasoning-process features are particularly useful when difficulty depends on multi-step problem solving, while the improvement on SAT Reading \& Writing indicates that the proposed representation also generalizes to text-based assessment items. A similar trend appears in regression. Our method achieves the strongest results on both Cambridge and USMLE, outperforming the best baseline models in terms of both error reduction and goodness of fit. In contrast, several prompting-based and LLM fine-tuning baselines perform poorly on these regression tasks, with some even producing negative $R^2$ values. This suggests that our method is especially beneficial for continuous difficulty estimation.

Overall, the results show that simply scaling model capacity does not reliably improve item difficulty prediction. By contrast, combining reasoning-trace-based process features with semantic item representations yields consistent gains across task types and domains, supporting our hypothesis that item difficulty is better captured by the structure of the underlying problem-solving process than by item text or direct model judgments alone.

\begin{table*}[t]
\centering
\small
\setlength{\tabcolsep}{4pt}
\resizebox{0.9\textwidth}{!}{
\begin{tabular}{lcc|cc|cc|cc|cc|cc}
\toprule
& \multicolumn{2}{c}{\textbf{QwQ}}
& \multicolumn{2}{c}{\textbf{Qwen}}
& \multicolumn{2}{c}{\textbf{Both}}
& \multicolumn{2}{c}{\textbf{QwQ (Role)}}
& \multicolumn{2}{c}{\textbf{Qwen (Role)}}
& \multicolumn{2}{c}{\textbf{Both (Role)}} \\
\cmidrule(lr){2-3}
\cmidrule(lr){4-5}
\cmidrule(lr){6-7}
\cmidrule(lr){8-9}
\cmidrule(lr){10-11}
\cmidrule(lr){12-13}
\textbf{Method} & \textbf{ACC} & \textbf{F1} & \textbf{ACC} & \textbf{F1} & \textbf{ACC} & \textbf{F1} & \textbf{ACC} & \textbf{F1} & \textbf{ACC} & \textbf{F1} & \textbf{ACC} & \textbf{F1} \\
\midrule
Sem. + FFea.        & \textbf{0.684} & \textbf{0.681} & \textbf{0.712} & \textbf{0.708} & \textbf{0.707} & \textbf{0.708} & \textbf{0.698} & \textbf{0.697} & \textbf{0.712} & \textbf{0.712} & \textbf{0.730} & \textbf{0.728} \\
\midrule
Sem. Only           & 0.609 & 0.607 & 0.609 & 0.607 & 0.609 & 0.607 & 0.609 & 0.607 & 0.609 & 0.607 & 0.609 & 0.607 \\
FFea. Only          & 0.670 & 0.670 & 0.674 & 0.674 & 0.684 & 0.677 & 0.614 & 0.618 & 0.660 & 0.657 & 0.637 & 0.631 \\
Len. Only           & 0.540 & 0.539 & 0.502 & 0.493 & 0.553 & 0.555 & 0.530 & 0.529 & 0.558 & 0.553 & 0.553 & 0.551 \\
Epi. Only           & 0.581 & 0.580 & 0.614 & 0.609 & 0.633 & 0.624 & 0.591 & 0.592 & 0.637 & 0.636 & 0.628 & 0.621 \\
Tran. Only          & 0.628 & 0.624 & 0.595 & 0.575 & 0.647 & 0.639 & 0.577 & 0.568 & 0.619 & 0.616 & 0.614 & 0.604 \\
Sem. w/o Len.       & 0.642 & 0.642 & 0.643 & 0.647 & 0.674 & 0.670 & 0.660 & 0.663 & 0.665 & 0.669 & 0.702 & 0.703 \\
Sem. w/o Epi.       & 0.651 & 0.652 & 0.651 & 0.647 & 0.656 & 0.646 & 0.665 & 0.666 & 0.679 & 0.679 & 0.688 & 0.690 \\
Sem. w/o Tran.      & 0.647 & 0.647 & 0.679 & 0.676 & 0.684 & 0.684 & 0.670 & 0.673 & 0.688 & 0.689 & 0.674 & 0.676 \\
Sem. + Len.         & 0.637 & 0.635 & 0.656 & 0.652 & 0.670 & 0.664 & 0.660 & 0.660 & 0.679 & 0.678 & 0.670 & 0.671 \\
Sem. + Epi.         & 0.656 & 0.652 & 0.674 & 0.670 & 0.674 & 0.672 & 0.660 & 0.661 & 0.684 & 0.682 & 0.684 & 0.686 \\
Sem. + Tran.        & 0.642 & 0.641 & 0.665 & 0.663 & 0.660 & 0.656 & 0.647 & 0.644 & 0.688 & 0.688 & 0.698 & 0.693 \\
\bottomrule
\end{tabular}
}
\caption{\textbf{Ablation results on SAT Math} across reasoning-trace sources and feature combinations. \textit{QwQ} and \textit{Qwen} denote the reasoning model that generated the traces, while \textit{Both} denotes averaging features from both trace sources. \textit{Role} indicates averaging features across proficiency-conditioned model profiles. \textit{Sem.} denotes the semantic embedding, \textit{FFea.} denotes the full process-feature set comprising length, episode-distribution, and transition features, \textit{Len.} denotes the length feature, \textit{Epi.} denotes the episode-distribution feature, and \textit{Tran.} denotes the transition feature.
Additional results are in Table~\ref{tab:feature_ablation_sat_math_appdx}.
}
\label{tab:feature_ablation_sat_math}
\end{table*}

\subsection{Ablation Study}

To comprehensively evaluate our approach, we conduct an ablation study on the SAT Math dataset while keeping all other experimental settings fixed. The study examines two aspects: the aggregation strategy and the representation design. Results are reported in Table~\ref{tab:feature_ablation_sat_math}.

\paragraph{Aggregation strategy}
We evaluate increasingly rich aggregation settings, including single-source reasoning traces under direct prompting (\textit{QwQ} and \textit{Qwen}), source-level aggregation over the two trace generators (\textit{Both}), and profile-based aggregation, where traces from different proficiency-conditioned model profiles are aggregated either within a single source (\textit{QwQ (Role)} and \textit{Qwen (Role)}) or across both sources together (\textit{Both (Role)}). The effect of aggregation is not strictly monotonic, but richer aggregation tends to improve performance. For example, under the full representation (\textit{Sem. + FFea.}), source-level aggregation does not improve over the stronger single-source result: the macro-F1 of \textit{Both} is 0.708, matching \textit{Qwen} and exceeding \textit{QwQ} at 0.681. In contrast, profile-based aggregation is more beneficial when applied across both sources: \textit{Both (Role)} achieves the best overall performance, reaching 0.730 accuracy and 0.728 macro-F1, compared with 0.707/0.708 for \textit{Both}. These results suggest that jointly aggregating diverse model profiles and trace sources provides a richer characterization of the elicited reasoning processes.

A potential concern is that profile-based aggregation benefits simply from using more rollouts. We therefore compare profiled aggregation with a matched number of repeated direct-prompt generations in Appendix~\ref{app:matched_rollout_control}. Profiled aggregation remains stronger under the same rollout budget, particularly in the combined-source setting, where it achieves a macro-F1 of 0.728 compared with 0.714 for direct replication. This result suggests that the gain reflects structured profile diversity rather than additional stochastic samples alone.


\paragraph{Representation design.}
Across all aggregation settings, the full representation (\textit{Sem. + FFea.}) consistently achieves the best performance, showing the benefit of combining semantic and process-based features. Using either semantic embeddings alone or process features alone leads to clear drops, with \textit{Sem. Only} trailing the full model by 7.4--12.1 macro-F1 points. Removing any one process feature family also consistently hurts performance, indicating that length, episode-distribution, and transition features provide complementary signals. Their contributions vary by setting: length and episode features tend to matter more in simpler settings, while transition features are especially important in the richest aggregated setting, \textit{Both (Role)}. Although adding any single process feature family already improves over \textit{Sem. Only}, the full combination remains the most effective.
Notably, process features alone are competitive with semantic features in several settings, but the strongest results consistently require their combination.

\section{Interpreting Item Difficulty through Episode-induced Features}
\label{sec:interpreting_item_difficulty}

To better understand what episode-induced representations reveal about item difficulty beyond predictive accuracy, we use SAT Math as a testbed.
We ask whether these features add predictive signal beyond item semantics and trace-length features, and what directional difficulty patterns the fitted model associates with them.

\begin{findingbox}[title={Key Finding 1: Difficulty Is Associated with a Structured Process Signature.}]
Harder items are associated with more intermediate work, more implementation-centered effort, and stronger analysis, implementation transition signals in model-generated reasoning traces.
\end{findingbox}

Appendix~\ref{app:additional_importance_same}, Table~\ref{tab:standardized_ame_features} provides the main model-grounded evidence for this process signature.
The largest positive standardized marginal signals come from total trace length and the \textit{think} portion, while final \textit{answer} length has a much smaller positive signal.
This suggests that difficulty is reflected less in more verbose final responses than in additional intermediate problem-solving work.
At the episode level, \textit{Implement} has the strongest positive count signal, and its ratio signal is also positive.
In contrast, the ratios of \textit{Read}, \textit{Plan}, and \textit{Verify} are negative, suggesting that harder items are associated with a relative shift toward concrete solution execution rather than a proportional increase in all types of reasoning activity.

The transition signals suggest how this execution-centered burden is reflected in local trace dynamics.
The strongest positive transitions include \textit{Analyze}$\rightarrow$\textit{Implement}, \textit{Implement}$\rightarrow$\textit{Implement}, and \textit{Implement}$\rightarrow$\textit{Analyze}.
These patterns suggest that harder items are not only associated with more implementation overall, but also with more extended and iterative local execution: higher counts of these transitions are associated with higher fitted difficulty scores, indicating movement from analysis into implementation, sustained implementation across adjacent steps, and returns from implementation to analysis.
This is consistent with an analysis-implementation-refinement pattern, but we do not treat these transitions as direct evidence of a distinct verification phase.
Rather, the strongest transition-level predictive signals appear to come mainly from analysis--implementation dynamics, which helps explain why the standalone \textit{Verify} episode need not increase.

Together, these results give a compact interpretation of difficulty: length captures the \emph{scale} of the burden, episode composition captures its \emph{allocation}, and transitions capture its \emph{flow}.
The main insight is therefore not that difficult items simply make the model think longer, but that they are associated with a more structured and implementation-centered model reasoning process.

\begin{findingbox}[title={Key Finding 2: Episode Dynamics Provide Predictive Process Evidence.}]
Episode composition and transition dynamics contribute complementary process evidence for difficulty prediction.
\end{findingbox}

We next connect the structured-burden interpretation to the features used by the predictor.
We assess feature-group importance in the profile-aggregated \textit{Both (Role)} setting; Appendix~\ref{app:additional_importance_same}, Table~\ref{tab:feature_importance_appendix} reports the complete results across trace-source and profile-aggregation settings.
We interpret these feature groups as complementary views of item difficulty:
\textit{Sem.} captures item content, \textit{Len.} captures reasoning scale, \textit{Epi.} captures effort allocation, and \textit{Tran.} captures local reasoning flow.

The feature-importance analysis shows that semantic features remain an important anchor, especially under SHAP importance, as expected for item difficulty prediction.
At the same time, the combined episode-structural group, \textit{Epi.+Tran.}, is also informative.
In the \textit{Both (Role)} setting, \textit{Epi.+Tran.} reaches 0.136 in permutation importance, exceeding \textit{Sem.} at 0.084 and \textit{Len.} at 0.042.
Under SHAP importance, \textit{Epi.+Tran.} also receives a non-negligible value of 1.024, compared with 0.526 for \textit{Len.}, although \textit{Sem.} remains larger under this metric.
The complete results in Appendix~\ref{app:additional_importance_same}, Table~\ref{tab:feature_importance_appendix} show the same broad pattern, with variation in magnitude across trace-source and aggregation settings: episode-composition and transition features consistently contribute useful process-level information.

This importance pattern is consistent with the ablation results in Table~\ref{tab:feature_ablation_sat_math}.
The full representation achieves the strongest performance, and removing any process feature family weakens the model.
Together, the importance and ablation analyses suggest that Epi2Diff works by combining multiple forms of evidence rather than relying on a single proxy.
Semantic features describe \emph{what} the item is about, length features describe the \emph{scale} of reasoning it elicits, episode-composition features describe \emph{where} reasoning effort is allocated, and transition features describe \emph{how} the reasoning process unfolds.
This closes the loop with Finding 1: the same scale-allocation-flow structure that characterizes difficult items also provides a predictive signal for the final difficulty model.

\vspace{-0.5em}
\section{Conclusion}

We propose Epi2Diff, a framework that combines semantic item representations with episode-level features extracted from model-generated reasoning traces. By modeling reasoning scale, effort allocation, and local transitions alongside item content, the framework captures complementary signals of problem-solving burden. Across four benchmarks, it outperforms a broad range of classification and regression baselines. On SAT Math, higher difficulty is associated with more implementation-centered reasoning and stronger analysis-to-implementation transitions, rather than simply longer traces. These findings show that cognitively grounded abstractions of reasoning traces can provide scalable, interpretable proxy evidence for item difficulty modeling. Future work should test the robustness of these process signatures across additional assessment domains, reasoning models, and item formats.

\vspace{-0.5em}
\section*{Ethics and Broader Impact}
We use publicly available or previously released assessment data and collect no new human-subject data, private student records, or personally identifying information. The intended benefit is more interpretable evidence for item-difficulty modeling and item-quality review. However, automated predictions must not be the sole basis for test construction, item approval, placement, or licensure decisions. Any deployment should include expert review, validation on the target population, subgroup fairness analysis, and monitoring for distribution shift. Predictions may reflect biases in the source datasets, selected reasoning models, and episode-labeling pipeline, and may be unreliable outside the evaluated domains. Model-generated reasoning traces are proxy evidence, not direct observations of human cognition or substitutes for psychometric validation.

\clearpage
\newpage
\bibliographystyle{assets/plainnat}
\bibliography{main}

\clearpage
\newpage
\beginappendix

\section{Related Work}
\label{sec:detail_rw}

\subsection{Item Difficulty Prediction}
Estimating item difficulty has traditionally relied on response-data calibration under CTT or IRT, which remains central for calibrated assessment but is costly for newly authored items because it typically requires field testing before items can be operationalized~\citep{demars2010item,hsu2018automated,alkhuzaey2024text,peters2025text}.
Expert judgment has also been considered as a lower-cost alternative, but prior work reports limited or variable agreement between expert ratings and empirical psychometric difficulty estimates, limiting its reliability as a scalable substitute~\citep{conejo2014empirical,wauters2012item}.
To reduce this reliance on response data, early NLP work treated item difficulty prediction as a text-based supervised learning problem.
Early approaches also used artificial neural networks with linguistic, propositional, and cognitive-process features for reading-comprehension items~\citep{perkins1995predicting}, while later work extracted over 100 hand-crafted textual-complexity features, such as word length, syntactic complexity, and readability indices, to predict difficulty in language assessment~\citep{loukina2016textual}.
Subsequent work incorporated richer semantic representations and item-component similarities, including Word2Vec-based features~\citep{hsu2018automated}, and showed that NLP-derived linguistic and syntactic features are predictive for high-stakes medical multiple-choice questions~\citep{yaneva2019predicting}.
Neural models further extended this line, including CNN-based models for reading-comprehension difficulty prediction~\citep{he2021automatically}.
Overall, this literature has gradually shifted from hand-crafted linguistic and psychometric proxies toward learned textual representations~\citep{alkhuzaey2024text,peters2025text}.

More recently, transformer-based representations have become strong baselines for item difficulty prediction.
BERT features have been used in a multi-task generalized linear model to jump-start item-parameter estimates for adaptive language testing~\citep{mccarthy2021jump}, and Transformer-based models have been shown to outperform traditional feature-based, TF-IDF, and Word2Vec baselines for question difficulty estimation~\citep{benedetto2023quantitative,devlin2019bert}.
Other work combines statistical, machine-learning, and language-model approaches, while also showing that simpler feature-based models can remain competitive in some settings~\citep{tack2024itec}.
Most recently, comparisons across a broad family of SLM- and LLM-based approaches suggest that fine-tuned encoder models such as BERT and RoBERTa can outperform larger but less task-specialized LLM alternatives on a medical assessment benchmark~\citep{li2025item,liu2019roberta}.
In conclusion, these studies establish strong text-representation baselines for item difficulty prediction, but they largely model difficulty from item text, metadata, annotations, embeddings, or other item-level representations, leaving process-level solving burden less directly explored.

\subsection{LLM-Based Methods for Item Difficulty Prediction}
With the emergence of LLMs, recent work has explored using model outputs and behavior as additional signals for item difficulty prediction.
One line of work prompts LLMs to answer items, simulate examinee-like responses, generate rationales, or directly estimate difficulty from item content.
Zero-shot LLM-generated answers and answer variability have been used to support multiple-choice item difficulty and response-time prediction~\citep{rogoz2024unibucllm}.
Other work simulates medical-candidate responses across prompting scenarios and extracts features from these responses for downstream prediction~\citep{duenas2024upn}.
More broadly, strong item-solving ability does not necessarily translate into accurate estimates of human difficulty, suggesting that direct LLM judgments may reflect model-centric difficulty rather than empirical human difficulty~\citep{li2025can}.
Similarly, LLMs show only limited alignment with human-calibrated item discrimination, indicating that difficulty is not the only psychometric property that remains challenging for LLM-based assessment~\citep{chen2026llms}.
Another line of work uses LLMs less as end-to-end judges and more as generators of auxiliary predictive signals.
These signals include LLM-based pipelines using semantic and linguistic features such as named entities and semantic roles~\citep{veeramani2024large}, reasoning- and sampling-based features~\citep{feng2025reasoning}, uncertainty-aware features~\citep{zotos2024you}, direct ratings and LLM-extracted cognitive or linguistic features combined with tree-based machine-learning models~\citep{razavi2025estimating}, and fine-tuned SLM/LLM representations for item difficulty modeling~\citep{li2025item}.
These studies indicate that LLM outputs can provide useful auxiliary cues for item difficulty prediction, but they do not uniformly outperform strong specialized baselines, and their extracted signals are usually treated as predictive features rather than structured, interpretable evidence of solving processes aligned with human psychometric difficulty.

A related but distinct body of work studies difficulty in broader LLM learning and evaluation settings.
Prior studies examine how difficulty shapes easy-to-hard generalization, cross-difficulty transfer, and benchmark design where difficulty is scaled by required reasoning length~\citep{hase2024unreasonable,kordi2026revisiting,ding2025longreasonarena}.
This work is relevant because it shows that difficulty affects model behavior and evaluation, though it does not focus on empirically calibrated human item difficulty in educational assessment.

\subsection{Reasoning Traces as Process Signals}
A parallel line of research asks not whether LLMs can directly judge difficulty, but whether generated reasoning traces contain structured information about how models process problems.
Chain-of-thought prompting made intermediate reasoning steps explicit~\citep{wei2022chain}, and newer reasoning-oriented models further amplified this trend by producing longer, human-readable traces that may expose self-reflection, verification, and strategy adaptation during inference~\citep{guo2025deepseek}.
Recent work on mathematical reasoning and reasoning language models further situates such traces within broader efforts to analyze model reasoning behavior and organization~\citep{ahn2024large,besta2025reasoning}.
This suggests that reasoning traces may reflect model-side process structure beyond verbosity, although they should not be interpreted as direct observations of human cognition.

Recent work has started to operationalize this idea through more explicit structural abstractions.
For example, Schoenfeld's Episode Theory has been applied to reasoning-model traces, showing that model-generated solutions can be decomposed into interpretable functional episodes~\citep{li2025understanding}.
More generally, studies on reasoning efficiency and trace quality suggest that effective reasoning cannot be reduced to response length alone; rather, it depends on how models allocate effort across exploration, execution, review, and verification, and longer traces may contain redundant or inefficient reasoning~\citep{chen2024not,feng2025characterizes,bogdan2025thought,aggarwal2025l1,fan2025missing}.
Still, most prior trace-based work has focused on interpretability, behavioral diagnosis, or reasoning-efficiency analysis, rather than using reasoning traces as structured predictive signals for downstream human item difficulty estimation.
This leaves open the central question of our work: whether model-generated reasoning traces can be transformed into scalable and interpretable process-level signals for estimating the human difficulty of unseen items.


\section{Additional Details}
\subsection{Evaluation Metrics}
\label{appdx:detail_exp}

For classification tasks, we evaluate performance using accuracy and macro-F1. Given ground-truth labels $\{y_i\}_{i=1}^N$ and predictions $\{\hat{y}_i\}_{i=1}^N$, accuracy is defined as
\begin{equation}
\mathrm{ACC}=\frac{1}{N}\sum_{i=1}^{N}\mathbf{1}(\hat{y}_i=y_i),
\end{equation}
where $\mathbf{1}(\cdot)$ is the indicator function.

Macro-F1 is defined as the unweighted average of class-wise F1 scores:
\begin{equation}
\mathrm{Macro\text{-}F1}=\frac{1}{C}\sum_{c=1}^{C}F1_c,
\end{equation}
where $C$ is the number of classes. For each class $c$, the F1 score is computed as
\begin{equation}
F1_c=\frac{2P_cR_c}{P_c+R_c},
\end{equation}
where $P_c$ and $R_c$ denote the precision and recall for class $c$, respectively:
\begin{equation}
P_c=\frac{\mathrm{TP}_c}{\mathrm{TP}_c+\mathrm{FP}_c},
\qquad
R_c=\frac{\mathrm{TP}_c}{\mathrm{TP}_c+\mathrm{FN}_c}.
\end{equation}

For regression tasks, we report RMSE and $R^2$. Given ground-truth values $\{y_i\}_{i=1}^N$, predictions $\{\hat{y}_i\}_{i=1}^N$, and the mean target value
\begin{equation}
\bar{y}=\frac{1}{N}\sum_{i=1}^N y_i,
\end{equation}
RMSE is defined as
\begin{equation}
\mathrm{RMSE}=\sqrt{\frac{1}{N}\sum_{i=1}^{N}(\hat{y}_i-y_i)^2},
\end{equation}
and the coefficient of determination is defined as
\begin{equation}
R^2 = 1-\frac{\sum_{i=1}^{N}(y_i-\hat{y}_i)^2}{\sum_{i=1}^{N}(y_i-\bar{y})^2}.
\end{equation}
A lower RMSE indicates smaller prediction error, while a higher $R^2$ indicates better fit to the ground-truth difficulty values.

\subsection{Dataset Splits}
\label{app:dataset_splits}

For Cambridge, SAT Math, and SAT Reading \& Writing, we use fixed stratified 80/20 train--test splits with random seed 42. Table~\ref{tab:dataset_splits} reports the resulting number of items in each partition. For USMLE, we use the official train--test split provided with the dataset rather than creating a new split.

\begin{table}[t]
\centering
\small
\setlength{\tabcolsep}{12pt}
\renewcommand{\arraystretch}{1.1}
\begin{tabular}{lrrr}
\toprule
\textbf{Dataset} & \textbf{Train items} & \textbf{Test items} & \textbf{Total} \\
\midrule
Cambridge & 634 & 159 & 793 \\
SAT Math & 860 & 215 & 1,075 \\
SAT Reading \& Writing & 1,070 & 268 & 1,338 \\
\bottomrule
\end{tabular}
\caption{Numbers of items in the fixed train--test splits. Cambridge, SAT Math, and SAT Reading \& Writing use stratified 80/20 splits with seed 42. USMLE uses its official train--test split and is therefore not included in the table.}
\label{tab:dataset_splits}
\end{table}

\subsection{Feature-Group Importance}
\label{app:feature_importance_definition}

We quantify the contribution of each feature group using both group-wise permutation importance and SHAP importance on the held-out test set. Let $f$ denote the fitted predictor, let $(\mathbf{X}_{\mathrm{test}}, \mathbf{y}_{\mathrm{test}})$ denote the test set, and let $G$ be a group of feature columns. For SAT Math, the evaluation score $\mathcal{M}$ is macro-F1. We jointly permute all columns in $G$ using the same random permutation $\pi_G$, which preserves dependencies among features within the group. The permutation importance of $G$ is the resulting decrease in test performance:
\begin{equation}
I_{\mathrm{perm}}(G)=
\mathcal{M}\!\left(f,\mathbf{X}_{\mathrm{test}},\mathbf{y}_{\mathrm{test}}\right)
-
\mathcal{M}\!\left(f,\mathbf{X}_{\mathrm{test}}^{\pi_G},\mathbf{y}_{\mathrm{test}}\right).
\end{equation}
Thus, a larger value indicates that shuffling the group causes a larger degradation in predictive performance.

For SHAP importance, let $\phi_{ij}^{(c)}$ be the SHAP value of feature $j$ for test item $i$ and class $c$. For a feature group $G$, we report the sum of mean absolute SHAP values over its constituent features:
\begin{equation}
I_{\mathrm{SHAP}}(G)=
\sum_{j\in G}\frac{1}{N_{\mathrm{test}}}
\sum_{i=1}^{N_{\mathrm{test}}}
\left(\frac{1}{C}\sum_{c=1}^{C}\left|\phi_{ij}^{(c)}\right|\right),
\end{equation}
where $C$ is the number of classes. This class average is used for the multiclass SAT Math classifier; for a scalar-output model, the inner class average reduces to $|\phi_{ij}|$. We apply these definitions to the semantic (\textit{Sem.}), length (\textit{Len.}), episode-composition (\textit{Epi.}), transition (\textit{Tran.}), and combined episode-plus-transition (\textit{Epi.+Tran.}) groups reported in Section~\ref{sec:interpreting_item_difficulty} and Appendix~\ref{app:additional_importance_same}.

\clearpage

\section{Supplemental Ablation Study}

\subsection{Transition Feature Design}

\begin{table*}[h]
\centering
\small
\setlength{\tabcolsep}{4pt}
\resizebox{\textwidth}{!}{
\begin{tabular}{lcc|cc|cc|cc|cc|cc}
\toprule
& \multicolumn{2}{c}{\textbf{QwQ}} 
& \multicolumn{2}{c}{\textbf{Qwen}} 
& \multicolumn{2}{c}{\textbf{Both}} 
& \multicolumn{2}{c}{\textbf{QwQ (Role)}} 
& \multicolumn{2}{c}{\textbf{Qwen (Role)}} 
& \multicolumn{2}{c}{\textbf{Both (Role)}} \\
\cmidrule(lr){2-3}
\cmidrule(lr){4-5}
\cmidrule(lr){6-7}
\cmidrule(lr){8-9}
\cmidrule(lr){10-11}
\cmidrule(lr){12-13}
\textbf{Method} & \textbf{ACC} & \textbf{F1} & \textbf{ACC} & \textbf{F1} & \textbf{ACC} & \textbf{F1} & \textbf{ACC} & \textbf{F1} & \textbf{ACC} & \textbf{F1} & \textbf{ACC} & \textbf{F1} \\
\midrule
Base + 2-gram Cnt.                    & 0.684 & 0.681 & 0.712 & 0.708 & 0.707 & 0.708 & 0.698 & 0.697 & 0.712 & 0.712 & \textbf{0.730} & \textbf{0.728} \\
\midrule
Base + 2-gram Rat.                    & 0.670 & 0.667 & 0.684 & 0.679 & 0.674 & 0.670 & 0.688 & 0.687 & 0.702 & 0.700 & 0.688 & 0.690 \\
Base + 2-gram Unq.                    & 0.670 & 0.671 & 0.693 & 0.688 & 0.688 & 0.688 & 0.674 & 0.677 & 0.684 & 0.683 & 0.688 & 0.691 \\
Base + 3-gram Cnt.                    & 0.651 & 0.651 & 0.684 & 0.679 & 0.679 & 0.680 & 0.670 & 0.669 & 0.681 & 0.681 & 0.698 & 0.698 \\
Base + 3-gram Dedup.                  & 0.642 & 0.639 & 0.679 & 0.673 & 0.684 & 0.681 & 0.688 & 0.690 & 0.698 & 0.699 & 0.684 & 0.684 \\
\midrule
Base + 2-gram Cnt. + 2-gram Rat.      & 0.660 & 0.658 & 0.693 & 0.687 & 0.679 & 0.675 & 0.660 & 0.662 & 0.702 & 0.701 & 0.688 & 0.690 \\
Base + 2-gram Cnt. + 2-gram Unq.      & 0.651 & 0.649 & 0.684 & 0.679 & 0.674 & 0.669 & 0.674 & 0.673 & 0.702 & 0.703 & 0.721 & 0.721 \\
Base + 2-gram Cnt. + 3-gram Cnt.      & 0.647 & 0.645 & 0.698 & 0.693 & 0.698 & 0.697 & 0.679 & 0.680 & 0.679 & 0.682 & 0.688 & 0.688 \\
Base + 2-gram Cnt. + 3-gram Dedup.    & 0.651 & 0.646 & 0.665 & 0.656 & 0.665 & 0.660 & 0.679 & 0.679 & 0.698 & 0.699 & 0.684 & 0.686 \\
\bottomrule
\end{tabular}
}
\caption{
\textbf{Transition-design ablation on SAT Math across reasoning-trace sources and aggregation settings.}
\emph{Base} denotes the semantic representation combined with the non-transition process features, i.e., length and episode-composition features.
\emph{Cnt.}, \emph{Rat.}, and \emph{Unq.} denote count, global ratio, and unique-count variants, respectively.
\emph{Dedup.} denotes trigram transitions computed after collapsing consecutive repeated episode labels.
Across all settings, the default \emph{2-gram Cnt.} transition design performs best, while alternative parameterizations and concatenated variants do not yield consistent gains.
}
\label{tab:transition_variant_sat_math}
\end{table*}

We explored several alternative parameterizations of transition features rather than assuming from the outset that adjacent bigram counts would be the best choice. Specifically, in addition to the default raw bigram transition counts, we considered globally normalized bigram ratios, unique bigram counts, raw trigram counts, and deduplicated trigram counts computed after collapsing consecutive repeated episode labels, inspired by ThinkARM~\citep{li2025schoenfeld}. Empirically, however, none of these alternatives produced more consistent gains than the raw bigram count representation across our SAT Math settings. A plausible explanation is that adjacent bigram counts already capture the dominant local reasoning flow that is most relevant for difficulty prediction, while the alternative designs either discard useful magnitude information (e.g., ratio or unique-count variants) or move to a more specific and sparser representation (e.g., trigram variants).

We further tested whether alternative transition parameterizations could provide complementary information on top of the default bigram count features by concatenating them with the base representation. Some of these combinations remained competitive, but none consistently surpassed the default bigram count design. We therefore do not adopt mixed transition representations in the final model. One likely reason is that these variants are not sufficiently independent from the adjacent bigram counts: higher-order or reformulated transitions partially re-encode the same local-order signal, so concatenation mainly increases dimensionality, sparsity, and redundancy. This issue is especially pronounced for trigram features: with eight episode labels, the transition space expands from \(8^2=64\) bigram features to \(8^3=512\) trigram features, making the representation much sparser and more prone to amplifying noise under our relatively limited data regime. Taken together, these results suggest that raw adjacent bigram counts provide the best trade-off among information retention, robustness, compactness, and interpretability, which is why we use them as the default transition representation.

\subsection{Additional Feature-Importance and Directional-Effect Analyses}
\label{app:additional_importance_same}

This appendix provides additional analyses supporting the interpretation of SAT Math difficulty as structured problem-solving burden.
Table~\ref{tab:feature_importance_appendix} reports feature-group importance across reasoning-trace settings, while Table~\ref{tab:standardized_ame_features} reports fine-grained standardized average marginal effects (SAMEs) for individual process features under the profile-aggregated \textit{Both (Role)} setting.
Together, the two tables separate two complementary questions: whether each feature group contributes to prediction, and how individual process features directionally relate to the fitted difficulty score.

\begin{table*}[h]
\centering
\small
\setlength{\tabcolsep}{3pt}
\resizebox{0.95\textwidth}{!}{
\begin{tabular}{lcccccccc}
\toprule
\textbf{Feature}
& \textbf{QwQ}
& \textbf{Qwen}
& \textbf{QwQ$^{+}$}
& \textbf{Qwen$^{+}$}
& \textbf{Both}
& \textbf{QwQ (Role)}
& \textbf{Qwen (Role)}
& \textbf{Both (Role)} \\
\midrule
\multicolumn{9}{c}{\textit{Permutation Importance}} \\
\midrule
Sem.       & --     & --     & 0.130 & 0.058 & 0.095 & 0.116 & 0.103 & 0.084 \\
Len.       & 0.055  & 0.019 & 0.062 & 0.027 & 0.052 & 0.046 & 0.045 & 0.042 \\
Epi.       & 0.130  & 0.225  & 0.096 & 0.089 & 0.069 & 0.057 & 0.058 & 0.024 \\
Tran.      & 0.018  & 0.015  & 0.011 & 0.031 & 0.069 & 0.027 & 0.067 & 0.064 \\
Epi.+Tran. & 0.141  & 0.184  & 0.133 & 0.140 & 0.088 & 0.083 & 0.166 & 0.136 \\
\midrule
\multicolumn{9}{c}{\textit{SHAP Importance}} \\
\midrule
Sem.       & --     & --     & 3.380 & 2.420 & 4.683 & 4.175 & 3.775 & 4.273 \\
Len.       & 0.406  & 0.231  & 0.264 & 0.252 & 0.450 & 0.439 & 0.373 & 0.526 \\
Epi.       & 1.378  & 0.545  & 0.542 & 0.437 & 0.635 & 0.598 & 0.366 & 0.436 \\
Tran.      & 1.470  & 0.450  & 0.200 & 0.206 & 0.425 & 0.466 & 0.570 & 0.588 \\
Epi.+Tran. & 2.848  & 0.995  & 0.742 & 0.643 & 1.060 & 1.065 & 0.936 & 1.024 \\
\bottomrule
\end{tabular}
}
\caption{
\textbf{Feature importance on SAT Math across reasoning-trace settings.}
\textit{Sem.}, \textit{Len.}, \textit{Epi.}, and \textit{Tran.} denote the semantic, length, episode-composition, and transition features, respectively.
\textit{Epi.+Tran.} denotes the combined episode-composition and transition feature group.
\textit{QwQ} and \textit{Qwen} denote structure-only settings without semantic features, while the superscript $+$ in \textit{QwQ}$^{+}$ and \textit{Qwen}$^{+}$ indicates the corresponding full-feature settings that include semantic features.
\textit{Both} denotes the combined QwQ and Qwen trace-source setting, and \textit{Role} denotes averaging across direct and proficiency-conditioned weak, medium, and strong model profiles.
Permutation importance is computed by shuffling each feature as a group, and SHAP importance is reported as the mean absolute SHAP value aggregated at the level of \textit{Sem.}, \textit{Len.}, \textit{Epi.}, \textit{Tran.}, and \textit{Epi.+Tran.}
Semantic importance is not applicable to the structure-only settings \textit{QwQ} and \textit{Qwen}, so those entries are marked as ``--''.
}
\label{tab:feature_importance_appendix}
\end{table*}

Table~\ref{tab:feature_importance_appendix} shows that episode-induced process features provide predictive signal across settings.
Although semantic features remain a strong anchor when they are included, the combined episode-composition and transition group (\textit{Epi.+Tran.}) consistently contributes nonzero importance.
In several permutation-importance settings, \textit{Epi.+Tran.} is comparable to or larger than the semantic feature group.
For example, under Both (Role), \textit{Epi.+Tran.} reaches 0.136 in permutation importance, exceeding \textit{Sem.} at 0.084 and \textit{Len.} at 0.042.
The SHAP results lead to a compatible conclusion: under Both (Role), \textit{Epi.+Tran.} has a mean absolute SHAP importance of 1.024, larger than \textit{Len.} at 0.526.
These results suggest that the predictor uses not only item semantics and reasoning length, but also the organization of reasoning across episode types and transitions.

\begin{table*}[t]
\centering
\scriptsize
\setlength{\tabcolsep}{3pt}
\renewcommand{\arraystretch}{1.03}
\resizebox{\textwidth}{!}{%
\begin{tabular}{ll|ll|ll|ll}
\toprule
\multicolumn{2}{c|}{\textbf{Length}} & \multicolumn{2}{c|}{\textbf{Episode}} & \multicolumn{4}{c}{\textbf{Transition}} \\
\cmidrule(lr){1-2} \cmidrule(lr){3-4} \cmidrule(lr){5-8}
\textbf{Feature label} & \textbf{Std. AME} & \textbf{Feature label} & \textbf{Std. AME} & \textbf{Feature label} & \textbf{Std. AME} & \textbf{Feature label} & \textbf{Std. AME} \\
\midrule
Length (Total) & 0.2374 & Count (Imp) & 0.0875 & Trans (Ana $\rightarrow$ Imp) & 0.0917 & Trans (Mon $\rightarrow$ Ver) & 0.0000 \\
Length (Think) & 0.1509 & Count (Ana) & 0.0417 & Trans (Imp $\rightarrow$ Imp) & 0.0729 & Trans (Ana $\rightarrow$ Exp) & 0.0000 \\
Length (Answer) & 0.0383 & Ratio (Imp) & 0.0086 & Trans (Imp $\rightarrow$ Ana) & 0.0206 & Trans (Ans $\rightarrow$ Ana) & 0.0000 \\
 &  & Count (Plan) & 0.0073 & Trans (Ans $\rightarrow$ Ans) & 0.0105 & Trans (Read $\rightarrow$ Ans) & 0.0000 \\
 &  & Count (Mon) & 0.0053 & Trans (Plan $\rightarrow$ Ana) & 0.0074 & Trans (Exp $\rightarrow$ Ans) & 0.0000 \\
 &  & Count (Ans) & 0.0038 & Trans (Ana $\rightarrow$ Ana) & 0.0069 & Trans (Ana $\rightarrow$ Ver) & 0.0000 \\
 &  & Ratio (Ana) & 0.0024 & Trans (Imp $\rightarrow$ Plan) & 0.0059 & Trans (Ver $\rightarrow$ Read) & 0.0000 \\
 &  & Count (Read) & 0.0000 & Trans (Plan $\rightarrow$ Imp) & 0.0054 & Trans (Exp $\rightarrow$ Imp) & 0.0000 \\
 &  & Count (Exp) & -0.0003 & Trans (Ver $\rightarrow$ Imp) & 0.0052 & Trans (Read $\rightarrow$ Exp) & 0.0000 \\
 &  & Ratio (Ans) & -0.0012 & Trans (Imp $\rightarrow$ Read) & 0.0045 & Trans (Read $\rightarrow$ Imp) & 0.0000 \\
 &  & Ratio (Plan) & -0.0013 & Trans (Mon $\rightarrow$ Ana) & 0.0038 & Trans (Ver $\rightarrow$ Plan) & -0.0001 \\
 &  & Ratio (Read) & -0.0015 & Trans (Ana $\rightarrow$ Plan) & 0.0037 & Trans (Plan $\rightarrow$ Plan) & -0.0001 \\
 &  & Count (Ver) & -0.0017 & Trans (Mon $\rightarrow$ Mon) & 0.0019 & Trans (Ans $\rightarrow$ Mon) & -0.0001 \\
 &  & Ratio (Exp) & -0.0030 & Trans (Ans $\rightarrow$ Ver) & 0.0017 & Trans (Ver $\rightarrow$ Ana) & -0.0001 \\
 &  & Ratio (Mon) & -0.0192 & Trans (Ans $\rightarrow$ Read) & 0.0016 & Trans (Mon $\rightarrow$ Read) & -0.0001 \\
 &  & Ratio (Ver) & -0.0278 & Trans (Imp $\rightarrow$ Ans) & 0.0015 & Trans (Ana $\rightarrow$ Mon) & -0.0001 \\
 &  &  &  & Trans (Read $\rightarrow$ Mon) & 0.0011 & Trans (Read $\rightarrow$ Read) & -0.0001 \\
 &  &  &  & Trans (Imp $\rightarrow$ Mon) & 0.0009 & Trans (Exp $\rightarrow$ Read) & -0.0001 \\
 &  &  &  & Trans (Mon $\rightarrow$ Imp) & 0.0008 & Trans (Ver $\rightarrow$ Ver) & -0.0002 \\
 &  &  &  & Trans (Ana $\rightarrow$ Read) & 0.0005 & Trans (Read $\rightarrow$ Plan) & -0.0003 \\
 &  &  &  & Trans (Imp $\rightarrow$ Exp) & 0.0004 & Trans (Imp $\rightarrow$ Ver) & -0.0003 \\
 &  &  &  & Trans (Plan $\rightarrow$ Exp) & 0.0004 & Trans (Ans $\rightarrow$ Plan) & -0.0003 \\
 &  &  &  & Trans (Plan $\rightarrow$ Read) & 0.0004 & Trans (Ans $\rightarrow$ Exp) & -0.0003 \\
 &  &  &  & Trans (Plan $\rightarrow$ Ver) & 0.0004 & Trans (Exp $\rightarrow$ Plan) & -0.0003 \\
 &  &  &  & Trans (Ver $\rightarrow$ Ans) & 0.0004 & Trans (Exp $\rightarrow$ Ver) & -0.0004 \\
 &  &  &  & Trans (Mon $\rightarrow$ Ans) & 0.0003 & Trans (Exp $\rightarrow$ Mon) & -0.0004 \\
 &  &  &  & Trans (Mon $\rightarrow$ Exp) & 0.0003 & Trans (Exp $\rightarrow$ Exp) & -0.0005 \\
 &  &  &  & Trans (Plan $\rightarrow$ Mon) & 0.0003 & Trans (Ana $\rightarrow$ Ans) & -0.0007 \\
 &  &  &  & Trans (Read $\rightarrow$ Ana) & 0.0003 & Trans (Ver $\rightarrow$ Mon) & -0.0010 \\
 &  &  &  & Trans (Ans $\rightarrow$ Imp) & 0.0002 & Trans (Plan $\rightarrow$ Ans) & -0.0010 \\
 &  &  &  & Trans (Exp $\rightarrow$ Ana) & 0.0002 & Trans (Ver $\rightarrow$ Exp) & -0.0017 \\
 &  &  &  & Trans (Mon $\rightarrow$ Plan) & 0.0001 & Trans (Read $\rightarrow$ Ver) & -0.0029 \\
\bottomrule
\end{tabular}%
}
\caption{\textbf{Fine-grained standardized directional effects on SAT Math difficulty.}
Values are standardized average marginal effects (SAME) on the expected ordinal difficulty score of the fitted XGBoost classifier,
\(\hat{s}(x)=\hat{p}_{\mathrm{medium}}(x)+2\hat{p}_{\mathrm{hard}}(x)\),
where \(\mathrm{easy}=0\), \(\mathrm{medium}=1\), and \(\mathrm{hard}=2\).
Positive values indicate that increasing the corresponding feature by one empirical training-set standard deviation shifts the predicted distribution toward harder classes, whereas negative values indicate shifts toward easier classes.
The magnitude corresponds to the average change in expected difficulty score on the ordered three-class scale.
}
\label{tab:standardized_ame_features}
\end{table*}

Table~\ref{tab:standardized_ame_features} further explains the direction of these process signals.
Among length features, total trace length and \textit{think} length have the largest positive effects, whereas final \textit{answer} length has a smaller positive effect.
This indicates that higher predicted difficulty is associated mainly with additional intermediate reasoning rather than merely longer final responses.
At the episode level, \textit{Implement} has the largest positive count effect, and the \textit{Implement} ratio is also positive.
In contrast, the ratios of \textit{Verify}, \textit{Read}, and \textit{Plan} are negative.
This pattern suggests that difficulty is not simply associated with uniformly longer traces; instead, the reasoning budget shifts toward concrete solution execution.

The transition features provide a local sequential view of this execution-centered burden.
The strongest positive transition effect is \textit{Analyze}$\rightarrow$\textit{Implement}, followed by \textit{Implement}$\rightarrow$\textit{Implement} and \textit{Implement}$\rightarrow$\textit{Analyze}.
Positive effects for \textit{Plan}$\rightarrow$\textit{Implement}, \textit{Analyze}$\rightarrow$\textit{Analyze}, and \textit{Verify}$\rightarrow$\textit{Implement} further indicate that harder items tend to involve movement from analysis or planning into execution, sustained execution, and local refinement around implementation.
Therefore, the process signal captured by the episode-induced representation is not reducible to token count alone.
Length captures the scale of the induced burden, episode features capture how reasoning effort is allocated, and transition features capture how the reasoning process unfolds over time.

\subsection{Additional Ablation Results}

To provide a more comprehensive comparison with \ours, we further conduct several additional experiments with alternative feature configurations, as shown in Table~\ref{tab:feature_ablation_sat_math_appdx}. These comparisons include using only LLM-extracted item-text features, combining item semantic representations with LLM-extracted features, and using only embeddings derived from the reasoning trace. Overall, the results show that \ours remains competitive against these alternatives. The LLM-extracted features provide useful item-level signals, but relying on them alone or combining them with item semantics does not consistently improve over the proposed fine-grained feature design. In addition, the trace-only setting performs less favorably, suggesting that reasoning traces by themselves may not capture sufficient information for robust prediction. These findings further support the effectiveness of explicitly modeling fine-grained item features in \ours.

\begin{table*}[t]
\centering
\small
\setlength{\tabcolsep}{4pt}
\resizebox{0.9\textwidth}{!}{
\begin{tabular}{lcc|cc|cc|cc|cc|cc}
\toprule
& \multicolumn{2}{c}{\textbf{QwQ}} 
& \multicolumn{2}{c}{\textbf{Qwen}} 
& \multicolumn{2}{c}{\textbf{Both}} 
& \multicolumn{2}{c}{\textbf{QwQ (Role)}} 
& \multicolumn{2}{c}{\textbf{Qwen (Role)}} 
& \multicolumn{2}{c}{\textbf{Both (Role)}} \\
\cmidrule(lr){2-3}
\cmidrule(lr){4-5}
\cmidrule(lr){6-7}
\cmidrule(lr){8-9}
\cmidrule(lr){10-11}
\cmidrule(lr){12-13}
\textbf{Method} & \textbf{ACC} & \textbf{F1} & \textbf{ACC} & \textbf{F1} & \textbf{ACC} & \textbf{F1} & \textbf{ACC} & \textbf{F1} & \textbf{ACC} & \textbf{F1} & \textbf{ACC} & \textbf{F1} \\
\midrule
Sem. + FFea.        & \textbf{0.684} & \textbf{0.681} & \textbf{0.712} & \textbf{0.708} & \textbf{0.707} & \textbf{0.708} & \textbf{0.698} & \textbf{0.697} & \textbf{0.712} & \textbf{0.712} & \textbf{0.730} & \textbf{0.728} \\
\midrule
LLM Fea. Only 
& 0.674 & 0.673 & 0.674 & 0.674 & 0.669 & 0.668 
& 0.674 & 0.674 & 0.688 & 0.687 & 0.683 & 0.685 \\

Item Sem. + LLM Fea. 
& 0.655 & 0.654 & 0.655 & 0.653 & 0.660 & 0.658 
& 0.669 & 0.666 & 0.678 & 0.676 & 0.655 & 0.652 \\

Trace Emb. Only 
& 0.647 & 0.642 & 0.605 & 0.597 & 0.628 & 0.623 
& 0.628 & 0.622 & 0.637 & 0.632 & 0.656 & 0.653 \\
\bottomrule
\end{tabular}
}
\caption{Additional comparison results across alternative feature configurations. \textit{Sem. + FFea} denotes the original setting that combines item semantic representations with the proposed fine-grained features. \textit{LLM Fea. Only} uses only the features extracted by the LLM from the item text itself, following the LLM-based feature extraction strategy of~\citet{razavi2025estimating}. \textit{Sem. + LLM Fea.} combines item semantic representations with these LLM-extracted item-text features. \textit{Trace Emb. Only} uses only embeddings of the reasoning trace, without item semantic or feature-based inputs.}
\label{tab:feature_ablation_sat_math_appdx}
\end{table*}

\subsection{Matched Rollout-Count Control for Role-Based Aggregation}
\label{app:matched_rollout_control}

One potential concern is that the benefit of role-based aggregation may come simply from using more reasoning-model rollouts, rather than from the proficiency-conditioned solver profiles themselves.
To separate the effect of additional rollouts from the effect of profile conditioning, we conduct a matched rollout-count control experiment on SAT Math.

We compare three settings.
\textit{Single Direct} uses one direct-prompt reasoning trace from each reasoning model.
\textit{Direct Replication} uses four independent generations from the same direct prompt for each reasoning model.
\textit{Profiled Aggregation} uses four proficiency-conditioned solver profiles.
Thus, \textit{Direct Replication} and \textit{Profiled Aggregation} use the same number of reasoning-model rollouts, but only \textit{Profiled Aggregation} introduces proficiency-conditioned profiles.
This matched comparison tests whether the observed benefit comes from additional stochastic samples alone or from structured profile-based aggregation.

\begin{table}[t]
\centering
\small
\setlength{\tabcolsep}{8pt}
\renewcommand{\arraystretch}{1.08}
\begin{tabular}{llcc}
\toprule
\textbf{Trace source} & \textbf{Setting} & \textbf{ACC} & \textbf{Macro-F1} \\
\midrule
QwQ  & Single Direct        & 0.684 & 0.681 \\
QwQ  & Direct Replication   & 0.685 & 0.683 \\
QwQ  & Profiled Aggregation & \textbf{0.698} & \textbf{0.697} \\
\midrule
Qwen & Single Direct        & 0.712 & 0.708 \\
Qwen & Direct Replication   & 0.710 & 0.710 \\
Qwen & Profiled Aggregation & \textbf{0.712} & \textbf{0.712} \\
\midrule
Both & Single Direct        & 0.707 & 0.708 \\
Both & Direct Replication   & 0.712 & 0.714 \\
Both & Profiled Aggregation & \textbf{0.730} & \textbf{0.728} \\
\bottomrule
\end{tabular}
\caption{
Matched rollout-count control on SAT Math.
\textit{Single Direct} uses one direct-prompt rollout, \textit{Direct Replication} uses four independent generations from the same direct prompt, and \textit{Profiled Aggregation} uses four proficiency-conditioned solver profiles.
\textit{Direct Replication} and \textit{Profiled Aggregation} use the same number of reasoning-model rollouts, but only \textit{Profiled Aggregation} introduces proficiency-conditioned profiles.
}
\label{tab:matched_rollout_control}
\end{table}

Table~\ref{tab:matched_rollout_control} shows that simply increasing the number of stochastic direct-prompt rollouts does not explain the benefit of profile-based aggregation.
For QwQ, moving from \textit{Single Direct} to \textit{Direct Replication} yields only a small gain, from 0.681 to 0.683 macro-F1, whereas \textit{Profiled Aggregation} reaches 0.697.
For Qwen, \textit{Direct Replication} slightly improves macro-F1 over \textit{Single Direct}, but \textit{Profiled Aggregation} remains the best setting.
For the combined trace-source setting, \textit{Direct Replication} improves over \textit{Single Direct}, but still remains below \textit{Profiled Aggregation}, which achieves the best performance with 0.730 ACC and 0.728 macro-F1.

These results indicate that the gain from role-based aggregation is not merely an artifact of using more reasoning-model generations.
Rather, proficiency-conditioned profiles introduce structured variation in the induced reasoning traces, leading to a more robust characterization of item difficulty than repeated direct-prompt sampling alone.

\clearpage

\subsection{Representative Item-level Episode Profiles}
\label{app:rep_episode_cases}

\begin{table}[h]
\centering
\small
\renewcommand{\arraystretch}{0.95}
\begin{tabular}{lrrrrrrrr}
\toprule
ID & Diff. & Total & Think & Answer & \% Impl. & \% Anal. & \% Ver. & \% Read \\
\midrule
\# 217 & Easy   & 557 & 399 & 158 & 0.253 & 0.151 & 0.172 & 0.120 \\
\# 235 & Med & 795 & 618 & 177 & 0.347 & 0.189 & 0.156 & 0.079 \\
\# 607 & Hard   & 995 & 728 & 267 & 0.403 & 0.203 & 0.130 & 0.074 \\
\bottomrule
\end{tabular}
\caption{\textbf{Characterization of representative SAT Math reasoning traces generated by Qwen3-32B.} As difficulty increases, total reasoning length and \textit{Implement} share increase, while \textit{Verify} and \textit{Read} decrease in relative share; \textit{Analyze} also increases.}
\label{tab:rep_item_stats}
\end{table}

\begin{table*}[h]
\centering
\small
\setlength{\tabcolsep}{4pt}
\renewcommand{\arraystretch}{1.15}
\begin{tabular}{ll}
\toprule
Item & Prefix of collapsed episode trajectory (first 8 episodes) \\
\midrule
Easy (\# 217) &
\texttt{Monitor(3) $\rightarrow$ Read(16) $\rightarrow$ Monitor(13) $\rightarrow$ Read(13)} \\
&
$\rightarrow$ \texttt{Plan(34) $\rightarrow$ Implement(12) $\rightarrow$ Plan(4) $\rightarrow$ Verify(6)  \dots} \\
\addlinespace[2pt]

Med (\# 235) &
\texttt{Read(29) $\rightarrow$ Monitor(2) $\rightarrow$ Verify(1) $\rightarrow$ Analyze(16)} \\
&
$\rightarrow$ \texttt{Verify(10) $\rightarrow$ Plan(15) $\rightarrow$ Monitor(6) $\rightarrow$ Analyze(20) \dots} \\
\addlinespace[2pt]

Hard (\# 607) &
\texttt{Monitor(11) $\rightarrow$ Analyze(43) $\rightarrow$ Read(9) $\rightarrow$ Monitor(5)} \\
&
$\rightarrow$ \texttt{Analyze(53) $\rightarrow$ Plan(5) $\rightarrow$ Implement(14) $\rightarrow$ Plan(5) \dots} \\
\bottomrule
\end{tabular}
\caption{\textbf{Prefixes of the collapsed episode sequences derived from Qwen3-32B reasoning traces for three representative SAT Math items.} For each item, we show the first 8 episodes in temporal order after collapsing consecutive identical labels and summing their token counts, split across two lines for readability. The Hard example already allocates substantial early budget to \textit{Analyze}, whereas the Easy example remains more fragmented across monitoring, reading, planning, and short execution/checking steps.}
\label{tab:rep_item_sequence_prefix}
\end{table*}

\begin{figure*}[t]
\centering
\begin{tcolorbox}[
    width=\textwidth,
    colback=white,
    colframe=black!60,
    boxrule=0.6pt,
    arc=1.5mm,
    left=2mm,
    right=2mm,
    top=1.5mm,
    bottom=1.5mm
]

\textbf{Item}
\vspace{1mm}

\begin{tcolorbox}[
    colback=gray!5,
    colframe=black!25,
    boxrule=0.4pt,
    arc=1mm,
    left=1.5mm,
    right=1.5mm,
    top=1mm,
    bottom=1mm
]
What is the value of $f(10)$?

The function $f$ is defined by $f(x)=4x-3$.

\textbf{Options:}
\begin{itemize}[leftmargin=1.8em, itemsep=1pt, topsep=2pt, parsep=0pt]
    \item[(A)] $-30$
    \item[(B)] $37$
    \item[(C)] $40$
    \item[(D)] $43$
\end{itemize}
\end{tcolorbox}

\vspace{1.5mm}
\textbf{Model Think}
\vspace{1mm}

\begin{center}
\begin{minipage}{0.94\textwidth}
\begin{Verbatim}[fontsize=\scriptsize, breaklines=true, breakanywhere=true]
<think>\nOkay, let's see. I need to find the value of f(10) for the function f(x) = 4x - 3. Hmm, functions can sometimes be confusing, but I think this one is straightforward. So, the function is given as f(x) equals 4 times x minus 3. That means for any value of x, I just plug it into this equation. \n\nSo, if I want f(10), I should replace every x in the function with 10. Let me write that out. So f(10) would be 4 multiplied by 10, and then subtract 3. Let me calculate that. 4 times 10 is 40, right? Then subtract 3 from 40. That would be 40 minus 3, which is 37. \n\nWait, let me check again. Maybe I made a mistake. Let me do the multiplication first. 4*10 is definitely 40. Then subtract 3. Yeah, 40 - 3 is 37. So the answer should be 37. Let me look at the options. Option B is 37. So that's probably the correct answer.\n\nBut let me make sure I didn't mix up anything. The function is f(x) = 4x - 3, not 4(x - 3). If it were 4(x - 3), then plugging in 10 would be 4*(10 - 3) = 4*7 = 28. But that's not the case here. The original function is 4x minus 3, so order of operations is important here. There's no parentheses around the x - 3, so it's just 4 times x, minus 3. So yeah, 4*10 is 40, minus 3 is 37. \n\nI think that's right. Let me check once more. Maybe I can plug in another number to see if the function works as expected. For example, if x is 1, then f(1) should be 4*1 - 3 = 1. Let me see. 4 times 1 is 4, minus 3 is 1. That seems correct. If x is 2, 4*2 is 8, minus 3 is 5. So f(2) is 5. That makes sense. So applying the same logic, f(10) is 4*10 - 3 = 40 - 3 = 37. Yep, that's consistent. \n\nTherefore, the correct answer should be option B, which is 37. I don't see any mistakes in my reasoning here. The key was substituting x with 10 correctly and following the order of operations. Since there's no parentheses around the x in the original function, it's just multiplication and then subtraction. So all steps check out.\n</think>
\end{Verbatim}
\end{minipage}
\end{center}

\vspace{1.5mm}
\textbf{Model Answer}
\vspace{1mm}

\begin{center}
\begin{minipage}{0.94\textwidth}
\begin{Verbatim}[fontsize=\scriptsize, breaklines=true, breakanywhere=true]
To solve the problem, we are given a function and asked to evaluate it at a specific input.\n\n---\n\n### Step 1: Understand the Function\n\nThe function is defined as:\n\n$$\nf(x) = 4x - 3\n$$\n\nThis is a linear function where the output is calculated by multiplying the input $ x $ by 4 and then subtracting 3.\n\n---\n\n### Step 2: Substitute the Input\n\nWe are asked to find $ f(10) $. To do this, we substitute $ x = 10 $ into the function:\n\n$$\nf(10) = 4(10) - 3\n$$\n\n---\n\n### Step 3: Perform the Arithmetic\n\n1. Multiply:  \n   $$\n   4 \\times 10 = 40\n   $$\n\n2. Subtract:  \n   $$\n   40 - 3 = 37\n   $$\n\n---\n\n### Step 4: Match with the Correct Option\n\nThe result is 37. Looking at the answer choices:\n\n- (A) $-30$\n- (B) $37$\n- (C) $40$\n- (D) $43$\n\nThe correct answer is **(B) 37**.\n\n---\n\n### Final Answer:\n\n$$\n\\boxed{B}\n$$
\end{Verbatim}
\end{minipage}
\end{center}

\vspace{2mm}
\begin{tcolorbox}[
    colback=gray!6,
    colframe=black!20,
    boxrule=0.4pt,
    arc=1mm,
    left=1mm,
    right=1mm,
    top=0.8mm,
    bottom=0.8mm
]
\textbf{Gold Answer:} B \hspace{1.5em} \textbf{Difficulty:} Easy
\end{tcolorbox}

\end{tcolorbox}
\caption{SAT Math item \# 217 (Easy), showing the full Qwen3-32B-generated thinking trace, final answer, gold answer, and difficulty label.}
\label{fig:full_example_verbatim}
\end{figure*}

\begin{figure*}[t]
\centering
\begin{tcolorbox}[
    width=\textwidth,
    colback=white,
    colframe=black!60,
    boxrule=0.6pt,
    arc=1.5mm,
    left=2mm,
    right=2mm,
    top=1.5mm,
    bottom=1.5mm
]

\textbf{Item}
\vspace{1mm}

\begin{tcolorbox}[
    colback=gray!5,
    colframe=black!25,
    boxrule=0.4pt,
    arc=1mm,
    left=1.5mm,
    right=1.5mm,
    top=1mm,
    bottom=1mm
]
What is the slope of line $n$?

When line $n$ is graphed in the xy-plane, it has an x-intercept of $(-4,0)$ and a y-intercept of $(0, \frac{86}{3})$.

\textbf{Options:}
\begin{itemize}[leftmargin=1.8em, itemsep=1pt, topsep=2pt, parsep=0pt]
    \item[(A)] $\frac{3}{344}$
    \item[(B)] $\frac{6}{\frac{40}{3}}$
    \item[(C)] $\frac{43}{6}$
    \item[(D)] $\frac{344}{3}$
\end{itemize}
\end{tcolorbox}

\vspace{1.5mm}
\textbf{Model Think}
\vspace{1mm}

\begin{center}
\begin{minipage}{0.92\textwidth}
\begin{Verbatim}[fontsize=\tiny, breaklines=true, breakanywhere=true]
<think>\nOkay, so I need to find the slope of line n. The problem says that the line has an x-intercept at (-4, 0) and a y-intercept at (0, 86/3). Hmm, slope... right. The slope of a line can be found if you have two points on the line. Since they gave me both intercepts, those are two points. So I can use the formula for slope, which is (y2 - y1)/(x2 - x1). Let me recall that formula correctly. Yeah, slope is rise over run, so the difference in y divided by the difference in x between two points.\n\nAlright, let me label the points. Let's take the x-intercept as (-4, 0). Let me call that (x1, y1). And the y-intercept is (0, 86/3), so that would be (x2, y2). Wait, actually, does it matter which one I choose as first or second? I think not, as long as I'm consistent. But let me stick with this.\n\nSo plugging into the formula: slope m = (y2 - y1)/(x2 - x1). So substituting the values, y2 is 86/3, y1 is 0. So numerator is 86/3 - 0 = 86/3. Denominator is x2 - x1, which is 0 - (-4) = 0 + 4 = 4. So the slope is (86/3)/4. Dividing by 4 is the same as multiplying by 1/4, so 86/3 * 1/4 = 86/(12). Let me simplify that. 86 divided by 12... Let's see if they have a common factor. Both are even numbers, so divide numerator and denominator by 2: 43/6. So the slope is 43/6. Let me check the answer choices. Option C is 43/6. So that must be the answer. \n\nWait, let me double-check my steps. First, the two points are correct. The x-intercept is where y=0, so (-4,0), and the y-intercept is where x=0, so (0, 86/3). Correct. Then slope formula: change in y over change in x. From (-4,0) to (0, 86/3), the change in y is 86/3 - 0 = 86/3. The change in x is 0 - (-4) = 4. So slope is (86/3)/4. Yes, that's 86/12, simplifies to 43/6. Yep. So option C is correct. Let me check if any other options could be equivalent. Option B is 6 divided by (40/3)... Wait, that would be 6*(3/40)= 18/40 = 9/20, which is not 43/6. Option A is 3/344, which is way smaller. Option D is 344/3, which is much larger. So only option C matches. So I think that's correct. \n\nAlternatively, maybe I can think of the equation of the line. Since I have intercepts, the intercept form of a line is x/a + y/b = 1, where a is the x-intercept and b is the y-intercept. So here, a is -4? Wait, intercept form is usually for positive intercepts. Wait, actually, the intercept form is x divided by the x-intercept plus y divided by the y-intercept equals 1. But if the x-intercept is negative, does that affect it? Let me think. If the x-intercept is (-4, 0), then a is -4. So plugging into intercept form: x/(-4) + y/(86/3) = 1. Maybe that's more complicated. Alternatively, converting to slope-intercept form. Since the y-intercept is given as (0, 86/3), so in y = mx + b, b is 86/3. So if I can find m. Well, we already did that with the slope formula. But maybe using another method to confirm. Let's take the point (-4, 0) and plug into y = mx + b. So 0 = m*(-4) + 86/3. Solving for m: -4m = -86/3 => m = (-86/3)/(-4) = (86/3)/4 = same as before, 86/12 = 43/6. Yep, same answer. So definitely, the slope is 43/6, which is option C. \n\nTherefore, the correct answer is C.\n</think>
\end{Verbatim}
\end{minipage}
\end{center}

\vspace{1.5mm}
\textbf{Model Answer}
\vspace{1mm}

\begin{center}
\begin{minipage}{0.92\textwidth}
\begin{Verbatim}[fontsize=\tiny, breaklines=true, breakanywhere=true]
To determine the slope of line $ n $, we are given two key points on the line:\n\n- The **x-intercept** is $ (-4, 0) $\n- The **y-intercept** is $ \\left(0, \\frac{86}{3}\\right) $\n\n---\n\n### Step 1: Use the slope formula\n\nThe slope $ m $ of a line passing through two points $ (x_1, y_1) $ and $ (x_2, y_2) $ is given by:\n\n$$\nm = \\frac{y_2 - y_1}{x_2 - x_1}\n$$\n\nLet's label the points:\n\n- $ (x_1, y_1) = (-4, 0) $\n- $ (x_2, y_2) = \\left(0, \\frac{86}{3}\\right) $\n\nSubstitute into the formula:\n\n$$\nm = \\frac{\\frac{86}{3} - 0}{0 - (-4)} = \\frac{\\frac{86}{3}}{4}\n$$\n\n---\n\n### Step 2: Simplify the expression\n\n$$\n\\frac{\\frac{86}{3}}{4} = \\frac{86}{3} \\cdot \\frac{1}{4} = \\frac{86}{12}\n$$\n\nNow simplify $ \\frac{86}{12} $ by dividing both numerator and denominator by their greatest common divisor, which is 2:\n\n$$\n\\frac{86 \\div 2}{12 \\div 2} = \\frac{43}{6}\n$$\n\n---\n\n### Step 3: Match with the answer choices\n\nThe simplified slope is $ \\frac{43}{6} $, which matches **Option C**.\n\n---\n\n### Final Answer:\n\n$$\n\\boxed{\\text{C}}\n$$
\end{Verbatim}
\end{minipage}
\end{center}

\vspace{2mm}
\begin{tcolorbox}[
    colback=gray!6,
    colframe=black!20,
    boxrule=0.4pt,
    arc=1mm,
    left=1mm,
    right=1mm,
    top=0.8mm,
    bottom=0.8mm
]
\textbf{Gold Answer:} C \hspace{1.5em} \textbf{Difficulty:} Medium
\end{tcolorbox}

\end{tcolorbox}
\caption{SAT Math item \# 235 (Medium), showing the full Qwen3-32B-generated thinking trace, final answer, gold answer, and difficulty label.}
\label{fig:medium_example_q235}
\end{figure*}

\begin{figure*}[t]
\centering
\begin{tcolorbox}[
    width=\textwidth,
    colback=white,
    colframe=black!60,
    boxrule=0.6pt,
    arc=1.5mm,
    left=2mm,
    right=2mm,
    top=1.5mm,
    bottom=1.5mm
]

\textbf{Item}
\vspace{1mm}

\begin{tcolorbox}[
    colback=gray!5,
    colframe=black!25,
    boxrule=0.4pt,
    arc=1mm,
    left=1.5mm,
    right=1.5mm,
    top=1mm,
    bottom=1mm
]
What is the value of $a$?

Function $f$ is defined by $f(x)=-a^{x}+b$, where $a$ and $b$ are constants. In the xy-plane, the graph of $y=f(x)-12$ has a y-intercept at $(0,-\frac{75}{7})$. The product of $a$ and $b$ is $\frac{320}{7}$.
\end{tcolorbox}

\vspace{1.5mm}
\textbf{Model Think}
\vspace{1mm}

\begin{center}
\begin{minipage}{0.92\textwidth}
\begin{Verbatim}[fontsize=\tiny, breaklines=true, breakanywhere=true]
<think>\nOkay, let's try to figure out this problem step by step. So, the question is asking for the value of 'a' given some information about the function f(x) = -a^x + b. There are two pieces of information provided: first, the graph of y = f(x) - 12 has a y-intercept at (0, -75/7). Second, the product of a and b is 320/7. Alright, let's break this down.\n\nFirst, let me recall that the y-intercept of a graph occurs where x = 0. So, if the graph of y = f(x) - 12 has a y-intercept at (0, -75/7), that means when we plug in x = 0 into the function y = f(x) - 12, the result should be -75/7. Let me write that down.\n\nSo, y = f(0) - 12 = -75/7. Therefore, f(0) - 12 = -75/7. Let me solve for f(0). If I add 12 to both sides, f(0) = -75/7 + 12. Hmm, I need to compute that. Let me convert 12 to sevenths. 12 is 84/7, right? So, -75/7 + 84/7 is (84 - 75)/7 = 9/7. Therefore, f(0) = 9/7.\n\nNow, since f(x) is defined as -a^x + b, let me compute f(0). For any exponential function a^x, when x = 0, a^0 = 1, right? So, f(0) = -1 + b. Therefore, -1 + b = 9/7. Let me solve for b. Adding 1 to both sides gives b = 9/7 + 1. Again, converting 1 to sevenths: 7/7. So, 9/7 + 7/7 = 16/7. Therefore, b = 16/7.\n\nOkay, so now we have b. The problem also states that the product of a and b is 320/7. So, a * b = 320/7. Since we found that b is 16/7, we can substitute that into the equation. Let me write that:\n\na * (16/7) = 320/7.\n\nTo solve for a, divide both sides by (16/7). Dividing by a fraction is the same as multiplying by its reciprocal, so:\n\na = (320/7) / (16/7) = (320/7) * (7/16). The 7s cancel out, so we have 320/16. Let me compute that. 320 divided by 16. Well, 16 times 20 is 320, so 320/16 is 20. Therefore, a = 20. \n\nWait, that seems straightforward, but let me check my steps again to make sure I didn't make a mistake. Let's verify each step.\n\nFirst, starting with the y-intercept. The graph is y = f(x) - 12. At x = 0, y is -75/7. Therefore, f(0) - 12 = -75/7. Solving for f(0): f(0) = -75/7 + 12. Converting 12 to 84/7, so 84/7 - 75/7 = 9/7. That seems right. Then f(0) = -a^0 + b = -1 + b. So, -1 + b = 9/7. Therefore, b = 9/7 + 1. Which is 16/7. Correct. Then, a * b = 320/7. Substituting b = 16/7, so a = (320/7) / (16/7) = 20. That seems correct. \n\nBut let me check if there's any other condition or something I might have missed. The function is f(x) = -a^x + b. The problem didn't mention anything else about the graph, like another point or behavior, but we were given the product of a and b. So with the given information, we were able to find b first, then use the product to find a. \n\nIs there a possibility that a could be negative? Well, in exponential functions, the base a is typically positive and not equal to 1. Since it's an SAT problem, they probably expect a positive real number for a. Also, if a were negative, then a^x might not be a real number for all x, unless x is an integer. But since the problem doesn't specify any restrictions, and given that the answer comes out nicely as 20, which is positive, I think that's the correct answer.\n\nLet me just verify once again with the numbers. If a is 20 and b is 16/7, then their product is 20*(16/7) = 320/7, which matches the given condition. Then f(0) is -1 + 16/7 = 9/7, so y = f(x) -12 at x=0 is 9/7 - 12. Let me compute that again. 9/7 - 12 is 9/7 - 84/7 = -75/7, which is exactly the y-intercept given. So all the numbers check out. \n\nTherefore, I think the value of a is 20. \n\n**Final Answer**\nThe value of $ a $ is \\boxed{20}.\n</think>
\end{Verbatim}
\end{minipage}
\end{center}

\vspace{1.5mm}
\textbf{Model Answer}
\vspace{1mm}

\begin{center}
\begin{minipage}{0.92\textwidth}
\begin{Verbatim}[fontsize=\tiny, breaklines=true, breakanywhere=true]
To determine the value of $ a $, we are given the function:\n\n$$\nf(x) = -a^x + b\n$$\n\nand two key pieces of information:\n\n1. The graph of $ y = f(x) - 12 $ has a y-intercept at $ (0, -\\frac{75}{7}) $.\n2. The product $ ab = \\frac{320}{7} $.\n\n---\n\n### Step 1: Use the y-intercept to find $ f(0) $\n\nThe y-intercept occurs when $ x = 0 $. So, we substitute $ x = 0 $ into the expression $ y = f(x) - 12 $:\n\n$$\ny = f(0) - 12 = -\\frac{75}{7}\n$$\n\nSolving for $ f(0) $:\n\n$$\nf(0) = -\\frac{75}{7} + 12\n$$\n\nConvert 12 to a fraction with denominator 7:\n\n$$\n12 = \\frac{84}{7}\n$$\n\nSo:\n\n$$\nf(0) = \\frac{84 - 75}{7} = \\frac{9}{7}\n$$\n\n---\n\n### Step 2: Use the definition of $ f(x) $ to find $ b $\n\nRecall that:\n\n$$\nf(x) = -a^x + b\n$$\n\nSo:\n\n$$\nf(0) = -a^0 + b = -1 + b\n$$\n\nSet this equal to the value we just found:\n\n$$\n-1 + b = \\frac{9}{7}\n$$\n\nSolving for $ b $:\n\n$$\nb = \\frac{9}{7} + 1 = \\frac{9}{7} + \\frac{7}{7} = \\frac{16}{7}\n$$\n\n---\n\n### Step 3: Use the product $ ab = \\frac{320}{7} $ to find $ a $\n\nNow that we know $ b = \\frac{16}{7} $, substitute into the product:\n\n$$\na \\cdot \\frac{16}{7} = \\frac{320}{7}\n$$\n\nSolve for $ a $:\n\n$$\na = \\frac{320}{7} \\div \\frac{16}{7} = \\frac{320}{7} \\cdot \\frac{7}{16} = \\frac{320}{16} = 20\n$$\n\n---\n\n### Final Answer\n\n$$\n\\boxed{20}\n$$
\end{Verbatim}
\end{minipage}
\end{center}

\vspace{2mm}
\begin{tcolorbox}[
    colback=gray!6,
    colframe=black!20,
    boxrule=0.4pt,
    arc=1mm,
    left=1mm,
    right=1mm,
    top=0.8mm,
    bottom=0.8mm
]
\textbf{Gold Answer:} 20 \hspace{1.5em} \textbf{Difficulty:} Hard
\end{tcolorbox}

\end{tcolorbox}
\caption{SAT Math item \# 607 (Hard), showing the full Qwen3-32B-generated thinking trace, final answer, gold answer, and difficulty label.}
\label{fig:hard_example_q607}
\end{figure*}

\clearpage

\section{Additional Analysis Visualizations}
\label{app:add_visualizations}

In this appendix, we provide the full role-wise visualizations of the episode-induced process representation on SAT Math using Qwen3-32B generated reasoning traces. 
Each figure corresponds to one solver role (\textit{direct}, \textit{weak}, \textit{medium}, or \textit{strong}) and combines three complementary views: 
(1) the length summary across Easy, Medium, and Hard items, 
(2) the episode-ratio deltas between difficulty groups, and 
(3) the transition-delta heatmaps between difficulty groups.

\begin{figure*}[h]
    \centering
    \includegraphics[width=0.9\textwidth]{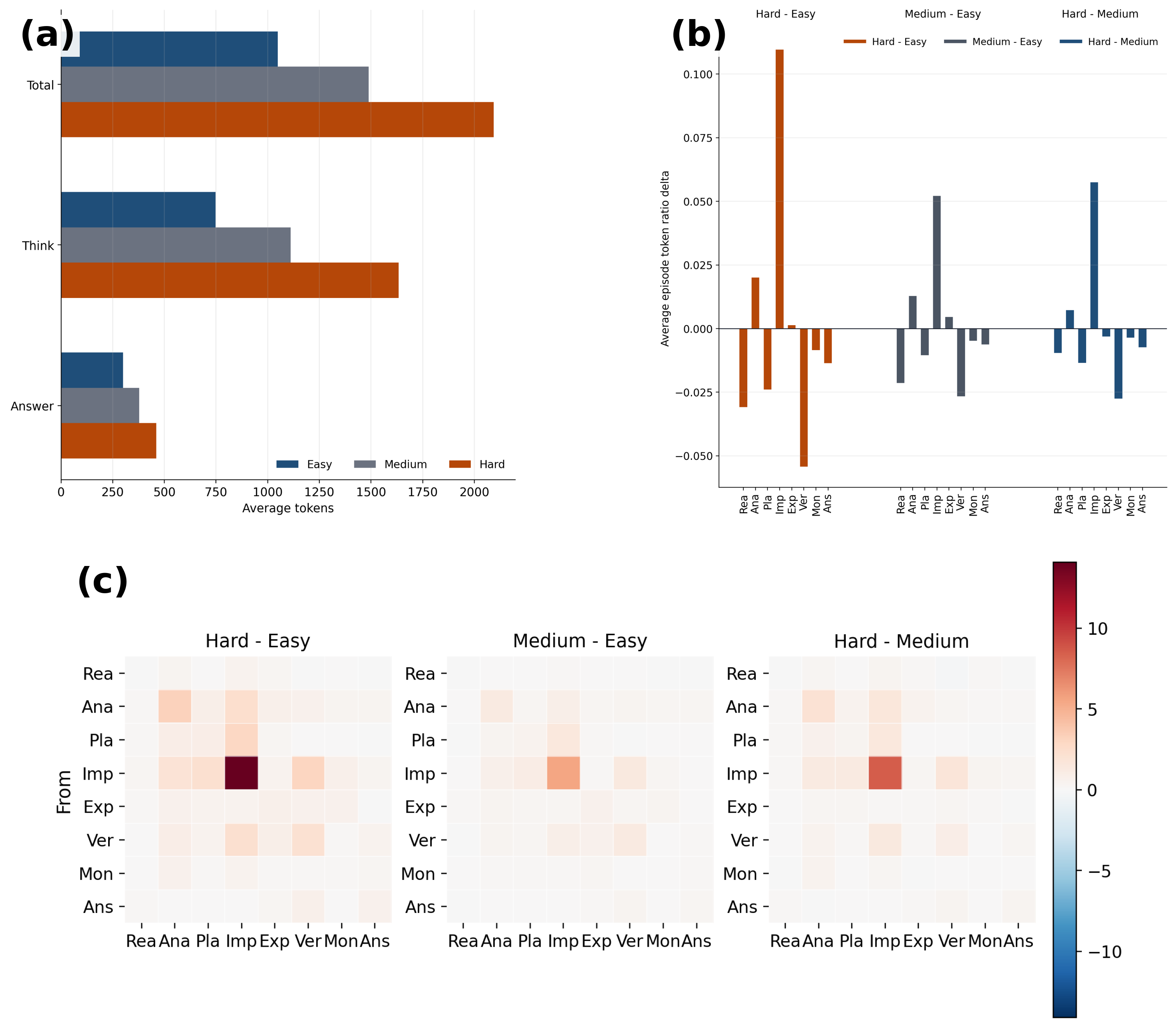}
    \caption{Role-wise visualization for the \textit{direct} solver profile on SAT Math. 
    The figure combines the length summary, episode-ratio deltas, and transition-delta heatmaps.} 
    \label{fig:direct_combined}
\end{figure*}

\begin{figure*}[h]
    \centering
    \includegraphics[width=0.9\textwidth]{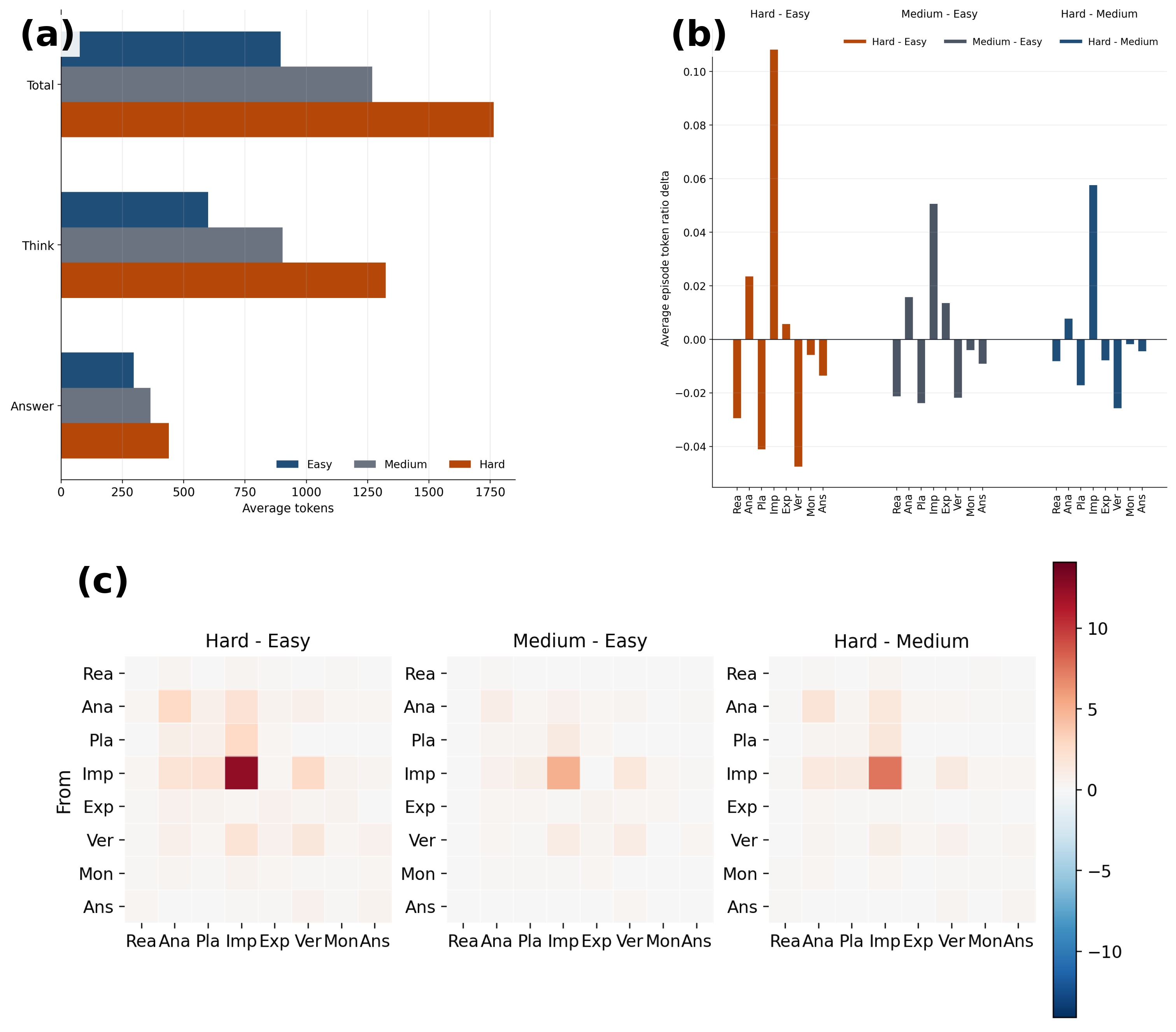}
    \caption{Role-wise visualization for the \textit{weak} solver profile on SAT Math. 
    The figure combines the length summary, episode-ratio deltas, and transition-delta heatmaps.} 
    \label{fig:weak_combined}
\end{figure*}

\begin{figure*}[h]
    \centering
    \includegraphics[width=0.9\textwidth]{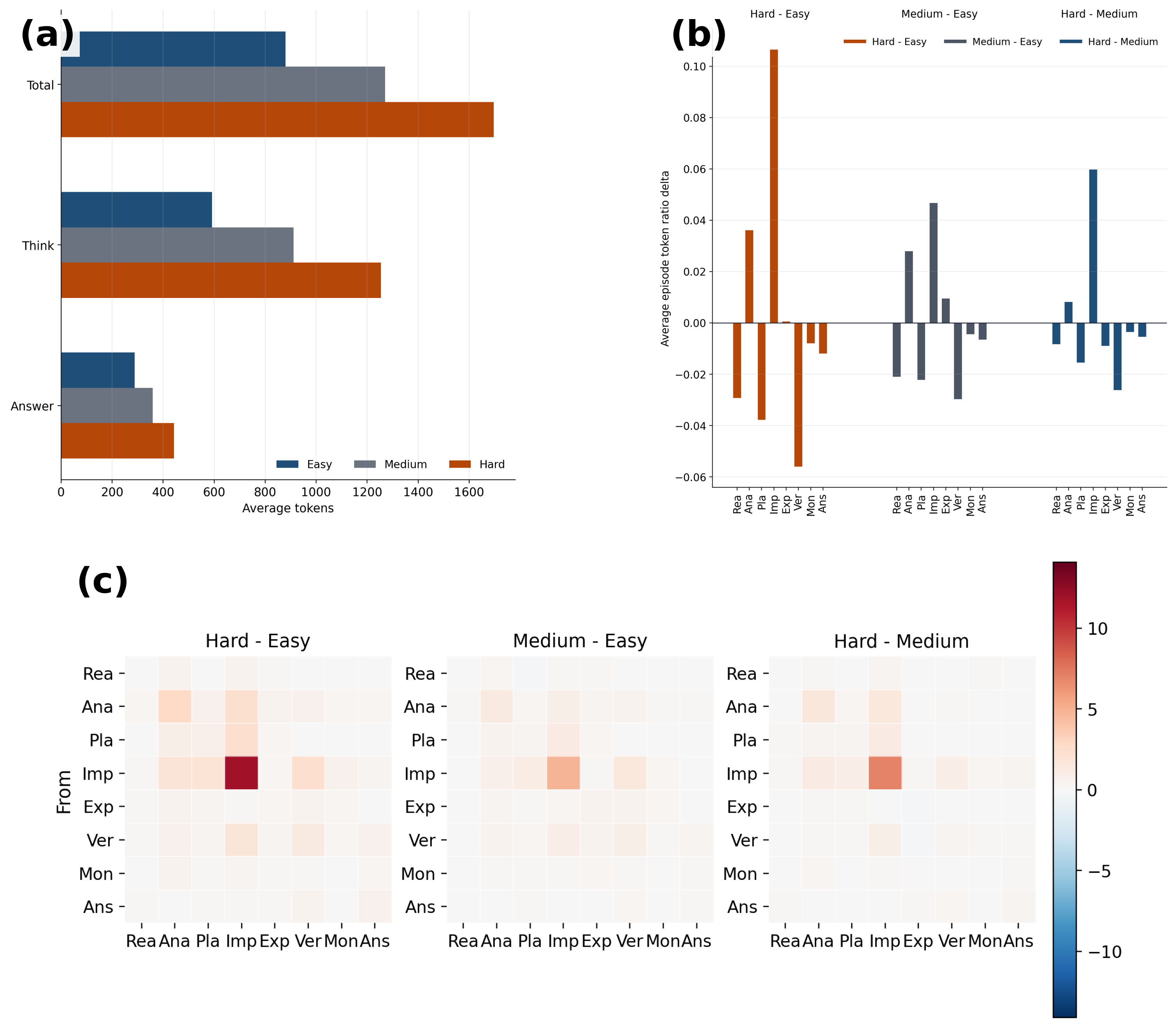}
    \caption{Role-wise visualization for the \textit{medium} solver profile on SAT Math. 
    The figure combines the length summary, episode-ratio deltas, and transition-delta heatmaps.} 
    \label{fig:medium_combined}
\end{figure*}

\begin{figure*}[h]
    \centering
    \includegraphics[width=0.9\textwidth]{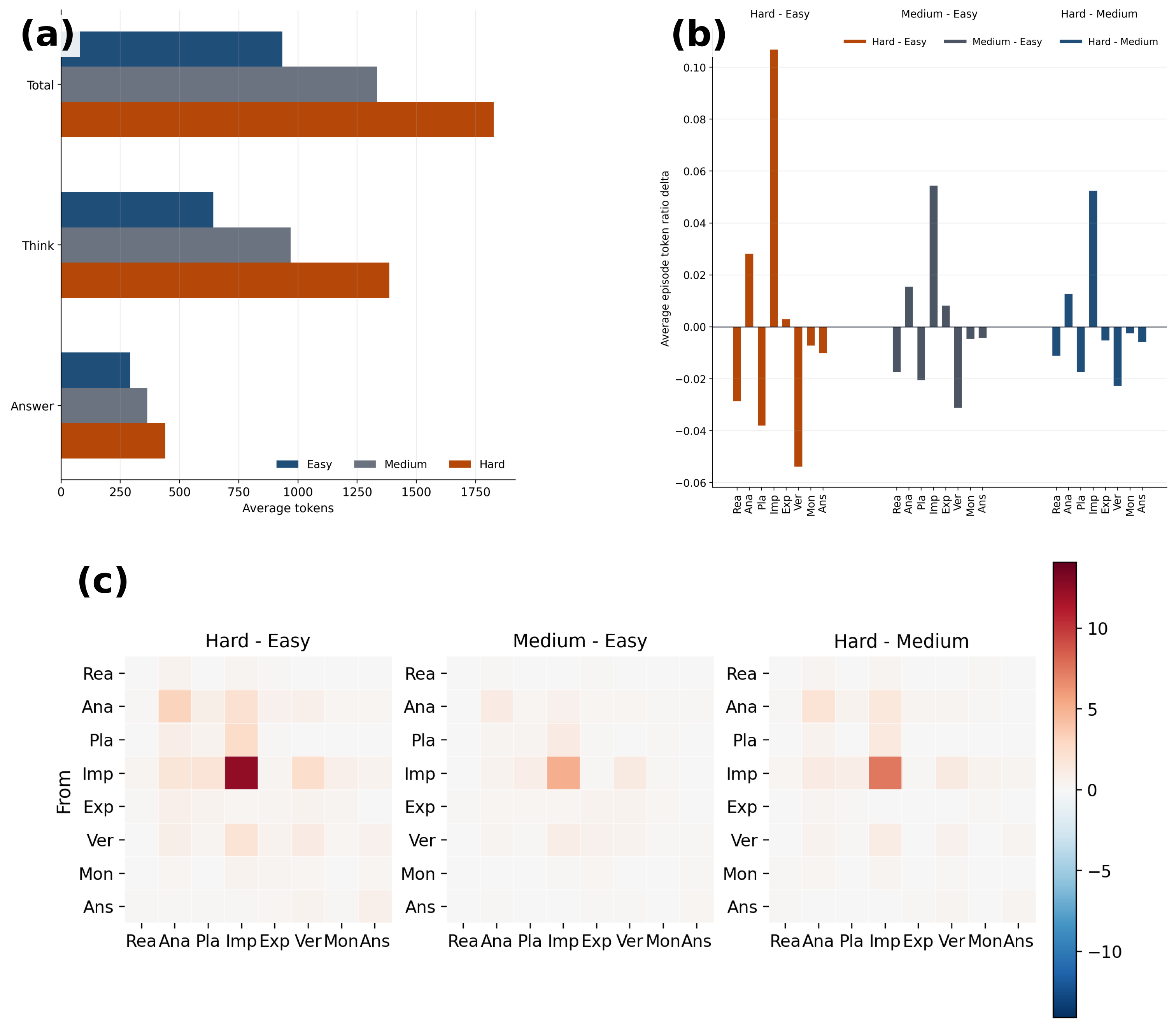}
    \caption{Role-wise visualization for the \textit{strong} solver profile on SAT Math. 
    The figure combines the length summary, episode-ratio deltas, and transition-delta heatmaps.}
    \label{fig:strong_combined}
\end{figure*}

\clearpage

\end{document}